\documentclass{article}

\PassOptionsToPackage{numbers}{natbib}

\usepackage[final]{neurips_2023}

\usepackage[utf8]{inputenc} %
\usepackage[T1]{fontenc}    %
\usepackage{hyperref}       %
\usepackage{caption}
\usepackage{url}            %
\usepackage{booktabs}       %
\usepackage{amsfonts}       %
\usepackage{nicefrac}       %
\usepackage{microtype}      %
\usepackage{xcolor}         %
\usepackage{multirow}
\usepackage{natbib}
\usepackage{graphicx}
\usepackage{footnote}
\usepackage{enumitem}
\makesavenoteenv{tabular}
\makesavenoteenv{table}

\title{Deductive Verification of Chain-of-Thought Reasoning}

\newcommand{\red}[1]{\textcolor{red}{#1}}

\newcommand{\prompt}[1][]{Natural Program}
\newcommand\blfootnote[1]{%
  \begingroup
  \renewcommand\thefootnote{}\footnote{#1}%
  \addtocounter{footnote}{-1}%
  \endgroup
}

\author{%
  Zhan Ling\textsuperscript{1}\thanks{Equal contribution}\quad Yunhao Fang\textsuperscript{1}\footnotemark[1]\quad Xuanlin Li\textsuperscript{1}\quad Zhiao Huang\textsuperscript{1}\quad Mingu Lee\textsuperscript{2}\\\textbf{Roland Memisevic\textsuperscript{2}\quad Hao Su\textsuperscript{1}}\\
  \textsuperscript{1}UC San Diego, \textsuperscript{2}Qualcomm AI Research\thanks{Qualcomm AI Research is an initiative of Qualcomm Technologies, Inc}
}
\begin{document}
\maketitle
\begin{abstract}
Large Language Models (LLMs) significantly benefit from Chain-of-Thought (CoT) prompting in performing various reasoning tasks. While CoT allows models to produce more comprehensive reasoning processes, its emphasis on intermediate reasoning steps can inadvertently introduce hallucinations and accumulated errors, thereby limiting models' ability to solve complex reasoning tasks. Inspired by how humans engage in careful and meticulous deductive logical reasoning processes to solve tasks, we seek to enable language models to perform \textit{explicit and rigorous deductive reasoning}, and also ensure the \textit{trustworthiness} of their reasoning process through self-verification. However, directly verifying the validity of an entire deductive reasoning process is challenging, even with advanced models like ChatGPT. In light of this, we propose to decompose a reasoning verification process into a series of step-by-step subprocesses, each only receiving their necessary context and premises. To facilitate this procedure, we propose \textbf{Natural Program}, a \textit{natural language-based} deductive reasoning format. Our approach enables models to generate precise reasoning steps where subsequent steps are more rigorously grounded on prior steps. It also empowers language models to carry out reasoning self-verification in a \textit{step-by-step} manner. By integrating this verification process into each deductive reasoning stage, we significantly enhance the rigor and trustfulness of generated reasoning steps. Along this process, we also improve the answer correctness on complex reasoning tasks. Code will be released at \url{https://github.com/lz1oceani/verify_cot}.
\end{abstract}

\section{Introduction}
The transformative power of large language models, enhanced by Chain-of-Thought (CoT) prompting~\cite{wei2022chain, kojima2022large, zhou2022least, shi2022language}, has significantly reshaped the landscape of information processing~\cite{drozdov2022compositional, lu2022learn, wei2022emergent, zeng2022socratic, driess2023palm, zelikman2022star, lampinen-etal-2022-language, marasovic2022fewshot}, fostering enhanced abilities across a myriad of disciplines and sectors. While CoT allows models to produce more comprehensive reasoning processes, its emphasis on intermediate reasoning steps can inadvertently introduce hallucinations~\cite{bubeck2023sparks, maynez2020faithfulness, guerreiro2023hallucinations, ji2023survey} and accumulated errors~\cite{bubeck2023sparks, welleck2019neural, arora2022exposure}, thereby limiting models' ability to produce cogent reasoning processes. 
\blfootnote{All datasets and models were solely downloaded and evaluated by the University of California San Diego.}

In fact, the pursuit of reliable reasoning is not a contemporary novelty; indeed, it is an intellectual endeavor that traces its roots back to the time of Aristotle's ancient Greece. Motivated by the desire to establish a rigorous reasoning process, in his ``Organon,'' Aristotle introduced principles of \emph{logic}, in particular, syllogism, a form of logical argument that applies deductive reasoning to arrive at a conclusion based on two or more propositions assumed to be true. In disciplines that rigorous reasoning is critical, such as judical reasoning and mathematical problem solving, documents must be written in a formal language with a logical structure to ensure the validity of the reasoning process. 

\begin{figure*}[t]
\centering
\includegraphics[width=\linewidth]{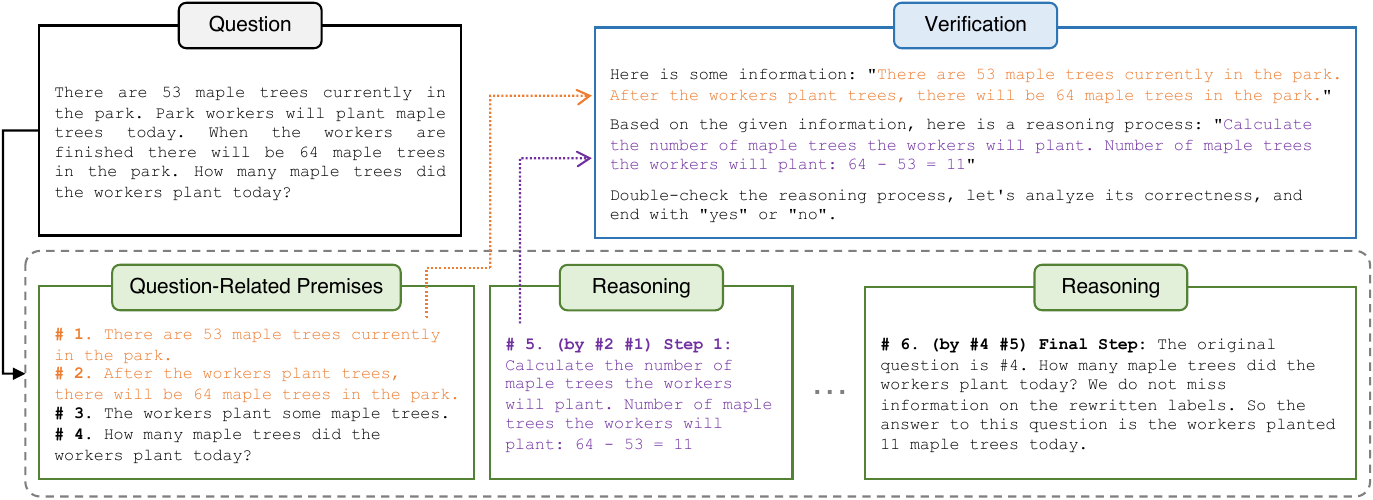}
\caption{An overview of our proposed deductive reasoning and verification process. In response to an input question, LLMs generate deductive reasoning chains using the \textit{Natural Program} format (bottom 3 boxes), a natural language-based deductive reasoning approach. The Natural Program format allows individual reasoning steps (an example in purple) and their corresponding minimal set of premises (an example in orange) to be easily extracted. This streamlined extraction process facilitates the step-by-step decomposition and verification of deductive reasoning (top-right box).}
\label{fig:natural_program}
\vspace{-1em}
\end{figure*}

\begin{figure*}[t]
\centering
\includegraphics[width=\linewidth]{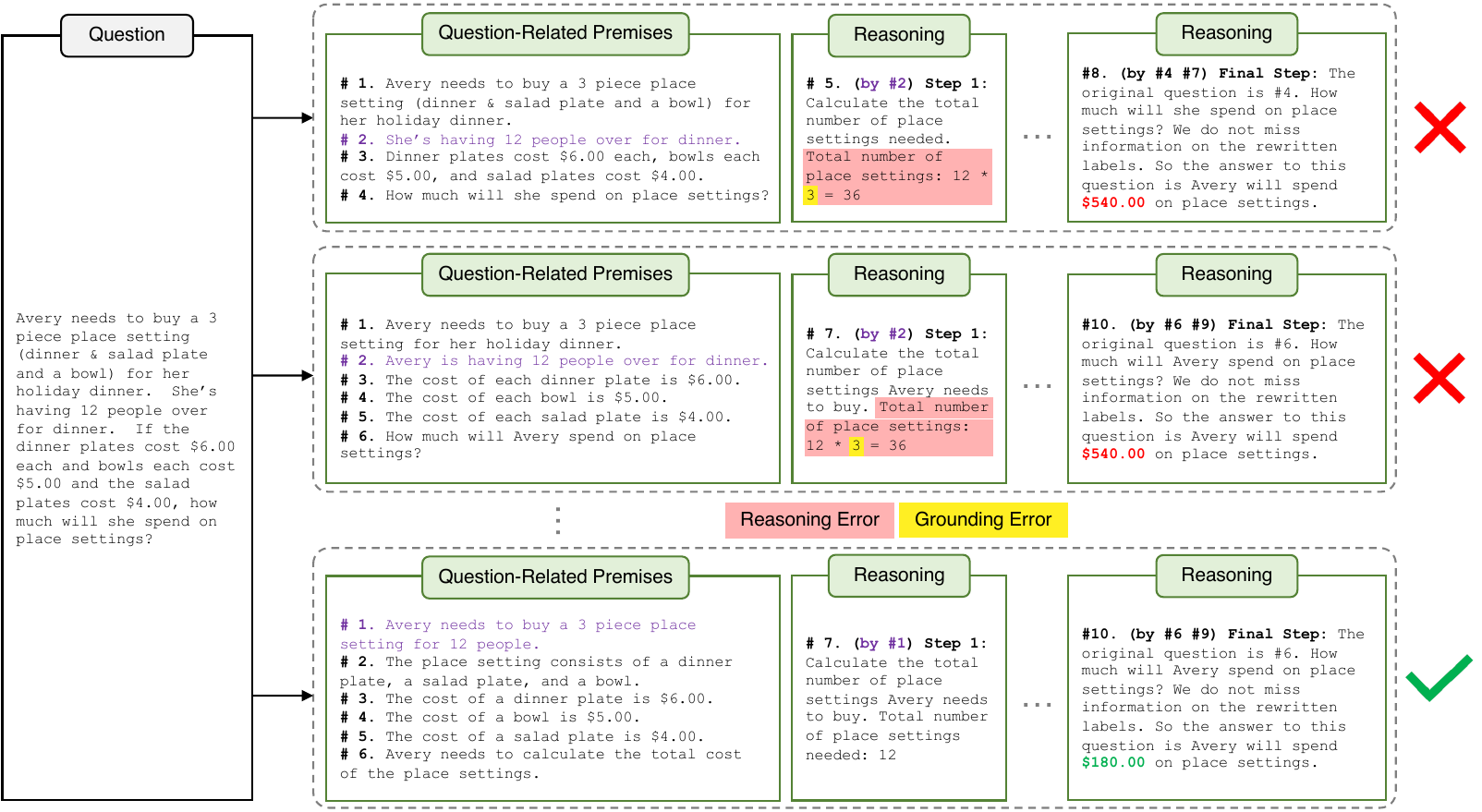}
\caption{Through our Natural Program-based deductive reasoning verification approach, we identify and eliminate reasoning chains that contain errors in reasoning and grounding (we define grounding error as utilizing information that is not present in cited premises). By alleviating such errors, we significantly enhance the rigor, trustworthiness, and interpretability of the generated reasoning outputs.}
\label{fig:natural_program2}
\vspace{-1em}
\end{figure*}

We yearn for this sequence of reliable knowledge when answering questions. Our goal is to develop language models that can propose potential solutions through reasoning in logical structures. Simultaneously, we aim to establish a verifier capable of accurately assessing the validity of these reasoning processes. Despite recent significant explorations in the field, such as \cite{wang2022self}'s emphasis on self-consistency and~\cite{lyu2023faithful, chen2022program}'s innovative use of codes to represent the reasoning process, these approaches still exhibit considerable limitations. For example, consistency and reliability are not inherently correlated; as for program codes, they are not powerful enough to represent many kinds of reasoning process, e.g., in the presence of quantifiers (``for all'', ``if there exists'') or nuances of natural language (moral reasoning, ``likely'', \ldots).         

We propose leveraging the power of natural language to achieve the deductive reasoning emphasized in ancient Greek logic, introducing a \textit{``natural program''}. This involves retaining natural language for its inherent power and avoiding the need for extensive retraining with large data sets. A natural program represents a rigorous reasoning sequence, akin to a computer program. We expect implementations of the idea to have two properties: 1) that natural programs are generated with minimal effort from an existing language model capable of CoT reasoning, preferably through in-context learning; 2) that the natural program can be easily verified for reliability in the reasoning process. 

Through a step-by-step investigation, we discovered that large language models have the potential to meet our expectation. Naïve CoT prompts like "Let us think step by step." has many flaws, and entrusting the entire verification process to a large model like ChatGPT can still lead to significant error rates. However, we found that, if the reasoning process is very short, and only based on necessary premises and contexts, the verification of existing large language models is already quite reliable. Therefore, our approach is to design prompts that induce CoT processes comprised of rigorous premises/conditions and conclusions with statement labels, and verification can be done by gradually isolating very few statements within the long thought chain. Experimentally, we found that most reasoning that passed the verification was rigorous, and many that did not pass had elements of imprecision in the reasoning process, even if they occasionally arrived at correct answers.

It is worth emphasizing that, we are not looking for a method to just maximize the correctness rate of final answers; instead, we aspire to generate a cogent reasoning process, which is more aligned with the spirit of judical reasoning. When combined with sampling-based methods, our method can identify low-probability but rigorous reasoning processes. When repeated sampling fails to yield a rigorous reasoning process, we can output "unknown" to prevent hallucinations that mislead users.

We demonstrate the efficacy of our natural program-based verification approach across a range of arithmetic and common sense datasets on publicly-available models like OpenAI's GPT-3.5-turbo.
Our key contributions are as follows:

1. We propose a novel framework for rigorous deductive reasoning by introducing a ``\textbf{Natural Program}'' format (Fig.~\ref{fig:natural_program}), which is suitable for verification and can be generated by just in-context learning;

2. We show that reliable self-verification of long deductive reasoning processes written in our Natural Program format can be achieved through step-by-step subprocesses that only cover necessary context and premises;

3. Experimentally, we demonstrate the superiority of our framework in improving the rigor, trustworthiness, and interpretability of LLM-generated reasoning steps and answers (Fig.~\ref{fig:natural_program2}).

\section{Related work}

\textbf{Reasoning with large language models.}
Recent large language models (LLMs)~\cite{brown2020language,chowdhery2022palm, zhang2022opt,touvron2023llama,scao2022bloom,hoffmann2022training,chung2022scaling,sanh2021multitask} have shown incredible ability in solving complex reasoning tasks. Instead of letting LLMs directly generate final answers as output, prior work have shown that by encouraging step-by-step reasoning through proper prompting, such as Chain-of-Thought (CoT) prompting~\cite{wei2022chain} and many others~\cite{kojima2022large,zhou2022least,zhang2022automatic,si2022prompting,wang2022self,zhou2022teaching,liu2023pre,yao2022react}, LLMs exhibit significantly better performance across diverse reasoning tasks. To further improve the step-by-step reasoning process, some recent studies have investigated leveraging external solvers such as program interpreters~\citep{schick2023toolformer,chen2022program, lyu2023faithful}, training and calling external reasoning modules~\citep{creswell2022faithful}, or performing explicit search to generate deductive steps~\cite{bostrom2022natural,proofwriter}. Parallel to these works, we do not rely on external modules and algorithms, and we directly leverage the in-context learning ability of LLMs to generate more precise and rigorous deductive reasonings. 
\vspace{3mm}

\textbf{Large language models as verifiers.} Using language models to evaluate model generations has been a long standing idea~\cite{kushman-etal-2014-learning,roy2016solving,shen2021generate, bubeck2023sparks}. As LLMs exhibit impressive capabilities across diverse tasks, it becomes a natural idea to use LLMs as evaluation and verification tools. For example, \cite{cobbe2021training, creswell2022faithful,paul2023refiner} finetune LLMs to verify solutions and intermediate steps. LLMs aligned with RLHF~\cite{ouyang2022training, openai2023gpt4, wang2022self} have also been employed to compare different model generations. In addition, recent works like \cite{shinn2023reflexion,weng2022verification,madaan2023self,chen2023teaching} leverage prompt designs to allow LLMs to self-verify, self-refine, and self-debug without the need for finetuning. However, these works do not focus on the rigor and trustworthiness of the deductive reasoning processes at every reasoning step. In this work, we propose a natural language-based deductive reasoning format that allows LLMs to self-verify \textit{every} intermediate step of a deductive reasoning process, thereby improving the rigor and trustfulness of reasoning.

Additionally, while some recent works~\cite{creswell2023selectioninference,yang2022nlproofs,golovneva2022roscoe,Prasad2023ReCEval} have proposed methods to verify individual steps in a reasoning process, our approach distinguishes from these works in the following perspectives: \textbf{(1)} Our approach leverages in-context learning to achieve reasoning verification, without the need for language model finetuning. \textbf{(2)} Our Natural Program-based LLM verification approach not only identifies invalid reasoning steps, but also provides explicit explanations for why they are invalid, detailing the specific reasoning errors involved. \textbf{(3)} Our Natural Program-based reasoning and verification approach is compatible with in-context abstract reasoning tasks where reasoning steps do not possess proof-like entailment structures. For example, our approach is compatible with the Last Letters task, where the LLM is instructed to output the concatenation of the last letters of all words in a sequence as the final answer. \textbf{(4)} Our Natural Program approach allows the use of commonsense knowledge not explicitly listed in premises. For example, consider this problem: ``Marin eats 4 apples a day. How many apples does he eat in November?'' Even though ``November has 30 days'' is not explicitly listed in the premises, Natural Program permits the use of such common knowledge within a reasoning step. Our in-context verification process is also capable of handling these implicit premises (e.g., if LLM outputs ``November has 29 days'' in a reasoning step, it will be marked as invalid).
\section{Motivation and Problem Formulation}
\label{sec:motivation}
A reasoning-based question-answering (QA) task can be defined as a tuple $(Q, C, O, A)$~\citep{ribeiro2023street}, where $Q$ is the target question; $C$ is the context of a question, such as the necessary background for answering a question; $O=(o_1, o_2, \cdots, c_k)$ are optional answer choices if $Q$ is a K-way multiple choice problem; and $A$ is the ground-truth answer. Given $Q$ and $C$ as inputs, large language models (LLMs)~\citep{brown2020language, chowdhery2022palm, touvron2023llama} generate a sequence of tokens $T=(t_1, t_2, \cdots, t_n)$ to answer the question. Recent works like Chain-of-Thought (CoT)~\citep{wei2022chain, kojima2022large} leverage prompt engineering in the context $C$ to encourage models to generate the intermediate reasoning process in $T$, which benefits LLM performance across diverse reasoning tasks. In this case, $T$ consists of a set of $m$ intermediate reasoning steps, which we denote as $S = (s_1, s_2, \cdots, s_m)$ . Each step $s_i$ can be represented by a subsequence of the generated tokens $(t_{l_i}, t_{r_i}) \subseteq T$. In much prior work, a generated solution is ``correct'' if and only if the predicted final answer in $s_m$ matches the ground truth $A$, which we call \textbf{answer correct(ness)}.

\newcommand{\wrongreasoning}[1]{\textcolor{red}{\textit{#1}}}
\newcommand{\correctreasoning}[1]{\textcolor{blue}{\textit{#1}}}

\begin{table}[t]
\centering
\resizebox{\textwidth}{!}{
\begin{tabular}{l}
\toprule
\textbf{Question}: Jerry is rolling a six-sided die. How much more likely is it (expressed as a percentage)\\ that he rolls a number greater than \$3 than that he rolls two even numbers in a row?\\
\textbf{Ground Truth}: There are 3 numbers greater than 3 on the dice, so the chances of rolling one \\ of them are \correctreasoning{3 / 6 = 50\%}. The chance of rolling one even number is 50\%, so the chance of rolling \\ two in a row is \correctreasoning{50\% * 50\% = 25\%}. The difference between these two probabilities is \correctreasoning{50\% - 25\%} \\\correctreasoning{ = 25\%}. \#\#\#\# 25\\
\midrule
\textbf{Output}: There are two possible outcomes for rolling a number greater than 3 out of 6 total \\outcomes, so the probability is \wrongreasoning{2 / 6 or 1 / 3}. For rolling two even numbers in a row, there are \\three possible outcomes (2-4, 4-6, 6-2) out of 36 total outcomes, so the probability is \wrongreasoning{3 / 36 or} \\ \wrongreasoning{1 / 12}. To find the difference in probability, we subtract the probability of rolling two even \\ numbers in a row from the probability of rolling a number greater than 3: \wrongreasoning{1 / 3 - 1 / 12 = 1 / 4}. \\ This means that it is 25\% more likely to roll a number greater than 3 than to roll two even \\ numbers in a row.
\\
\bottomrule
\end{tabular}
}
\vspace{0.5em}
\caption{An example question from GSM8K with a generated CoT reasoning path with GPT-3.5~(turbo), where the output provides a wrong reasoning chain with the correct answer.}
\label{tab:wrong_reasoning}
\vspace{-1em}
\end{table}

We observe that for all cases where LLMs produce erroneous final answers, there exists at least one mistake among the intermediate reasoning steps $S$. Moreover, even when the final answer is correct, there might still exist some mistakes among $S$. This phenomenon, as illustrated in Tab.~\ref{tab:wrong_reasoning}, occurs for all LLMs we tested, including state-of-the-art models such as ChatGPT and GPT-4~\cite{ouyang2022training}. Since later reasoning steps are conditioned on prior reasoning steps, these mistakes often initiate a snowball effect, causing subsequent mistakes to compound. This significantly diminishes the likelihood of correct problem-solving and impedes the progress towards achieving human-level complex reasoning.

 Therefore, in this work, we place significant emphasis on ensuring the validity of \textit{every} reasoning step, not just the correctness of the final answer. In particular, we focus on the validity of \textit{deductive reasoning}, an essential component of a logical reasoning process. In deductive reasoning, we are given a (premise, conclusion) pair, and we are interested in determining whether the conclusion follows from the premises. In the context of reasoning-based QA tasks, for each reasoning step $s_i$, we define its \textit{deductive validity} $V(s_i)$ as a binary variable. A reasoning step is \textbf{deductively valid} ($V(s_i)=1$) if and only if $s_i$ can be logically deduced from its corresponding premises $p_i$, which consist of the context $C$, the question $Q$, and all the previous reasoning steps $s_j (j < i)$. Then, we can also define the deductive validity for the entire reasoning chain $S$ as $V(S)=\wedge_{i=1}^M V(s_i)$. Compared to evaluating answer correctness, which can be accomplished by simple functions such as exact string match, evaluating deductive validity is a lot more challenging. Thanks to the recent progress on LLMs, which demonstrate impressive in-context learning capabilities across diverse scenarios, we propose to use LLMs to examine reasoning chains and predict the deductive reasoning validity.

\section{Deductively Verifiable Chain-of-Thought Reasoning}
\label{sec:main_method}

In this section, we introduce our specific approaches to performing deductive verification of reasoning chains. Specifically, we first introduce our motivation and method for decomposing a deductive verification process into a series of step-by-step processes, each only receiving contexts and premises that are necessary. Then, we propose \textbf{\prompt}, a natural language-based deductive reasoning format, to facilitate local step-by-step verification. Finally, we show that by integrating deductive verification with unanimity-plurality voting, we can improve the trustworthiness of reasoning processes along with final answers. An overview of our approach is illustrated in Fig.~\ref{fig:natural_program} and Fig.~\ref{fig:natural_program2}. %

\subsection{Decomposition of Deductive Verification Process}
\label{sec:method_decomposition}

\begin{table*}[t]
\centering
\small
\begin{tabular}{cccccccc} 
\toprule
Prompting & Reasoning Correctness & GSM8K & AQuA & MATH & AddSub & Date & Last Letters\\ 
\midrule
\multirow{3}{*}{Zero-shot}& 
Correct & 0.98 & 0.96 & 1.00 & 0.98 & 0.98 & 1.00\\
&Incorrect & 0.04 & 0.06 & 0.04 & 0.02 & 0.04 & 0.04\\
& (Average) & 0.51 & 0.51  & 0.52  &  0.50 &  0.51 & 0.52 \\
\midrule
\multirow{3}{*}{Two-shot} & Correct & 0.98 & 0.96 & 1.00 & 0.92 & 1.00 & 0.96\\
& Incorrect & 0.02 & 0.04 & 0.00 & 0.06 & 0.26 & 0.06\\
& (Average) & 0.50 & 0.50 & 0.50 & 0.49 & 0.63 & 0.51\\
\bottomrule
\end{tabular}
\caption{Zero-shot and two-shot reasoning chain verification accuracy for GPT-3.5-turbo (ChatGPT), where an entire reasoning chain is verified at once. The two shot prompt we used is presented in Appendix~\ref{app:no_natural_program_prompt}. To generate verification inputs, for each dataset, we perform Chain-of-Thought (CoT) prompting and randomly sample 50 reasoning chains that are valid and 50 reasoning chains that exhibit mistakes. We observe that when given an \textit{entire} reasoning process, where the deductive graphs for all reasoning steps are entangled together, it is challenging even for strong language models like ChatGPT to verify its validity.}
\label{tab:cot_verification}
\vspace{-1em}
\end{table*}

Given a reasoning chain $S=(s_1, s_2, \cdots, s_n)$, a straightforward idea to verify its deductive validity is to ask LLMs to examine the \textit{entire} reasoning chain at once. To assess the effectiveness of this approach, we conduct a preliminary experiment: for a dataset problem and its reasoning chain $S$ generated by ChatGPT, we prompt ChatGPT with ``\texttt{Do you think the above reasoning process is correct? Let's think step by step}'' such that its outputs whether there exists any mistake among any reasoning step in $S$. However, as demonstrated in Tab.~\ref{tab:cot_verification}, the verification accuracy is 50\% for most datasets, and ChatGPT struggles at finding out mistaken reasonings. Notably, it persistently outputs ``Correct'' for most reasoning chain queries, regardless of their actual validity.

We conjecture that such phenomenon is caused by the abundance of irrelevant premises for each reasoning step. Recall that the premises $p_i$ for a reasoning step $s_i$ consist of the the question $Q$, the question context $C$, along with the prior reasoning steps $s_{\le j}=\{s_j: j<i \}$. For $Q$ and $C$, we can further extract and decompose $Q\cup C$ into a set of ``question-related premises'' $QC=\{qc_1, qc_2, \cdots, qc_m\}$, where $qc_i$ is a premise or condition inferred from $Q\cup C$. Then, it is often the case that most elements of $p_i = QC \cup s_{\le j}$ are irrelevant to the validity of $s_i$, leading to erroneous verifications from language models. A very recent work~\cite{shi2023large} also observes a similar phenomenon where LLMs are easily distracted by irrelevant context.

Hence, we propose a decomposition of the reasoning chain verification process into a series of step-by-step processes, where each step only considers the premises that are \textit{necessary}. The overall validity of the reasoning chain, denoted as $V(S)=\wedge_{i=1}^M V(s_i)$, can be naturally decomposed into individual step validity ${V(s_i)}$. However, achieving such decomposition is highly challenging without imposing constraints on the format of reasoning chains. Additionally, for each $s_i \in S$, we aim to ensure that it \textit{explicitly} lists the minimal subset of premises $\bar{p_i} \subseteq p_i$ required for deductive reasoning to avoid potential ambiguities during verification. This motivates us to introduce a natural-language-based deductive reasoning format in Section \ref{sec:method_natural_program}.

\subsection{Natural Program Deductive Reasoning Format}
\label{sec:method_natural_program}
As previously mentioned in Sec.~\ref{sec:method_decomposition}, we desire LLMs to output deductive reasoning processes that can be easily verified by themselves, specifically by listing out the minimal set of necessary premises $p_i$ at each reasoning step $s_i$. To accomplish its goal, we propose to leverage the power of natural language, which is capable of rigorously representing a large variety of reasoning processes and can be generated with minimal effort. In particular, we introduce \textbf{\prompt}~, a novel deductive reasoning format for LLMs. More formally, \prompt~consists of the following components:

\begin{itemize}
    \item An instruction for models to extract question-related premises $QC$. We use the following instruction: ``\texttt{First, let's write down all the statements and relationships in the question with labels}".
    \item A numbered-list of question-related premises, each prefixed with ``\#\{premise\_number\}''.
    \item An instruction for models to generate the reasoning chain $S$ based on the question-related premises $QC$. We use the following instruction: ``\texttt{Next, let's answer the question step by step with reference to the question and reasoning process}''.
    \item A list of prefixed reasoning steps $S_i$. The prefix has the following format: \\ \texttt{\#\{number\} (by \{list\_of\_premises\_used\})}. Here ``number'' equals $|QC| + i$, and ``list\_of\_premises\_used'' consists of numbers from the smallest subset of premises among $QC\cup s_{\le j}$ that are used for the deductive reasoning of $s_i$. In addition, for the last reasoning step $s_m$, we ensure that it (1) includes a special tag \texttt{Final Step}; (2) refers to the premise number of the target question to be answered; (3) explicitly gives the final answer to a question.
\end{itemize}

To encourage language models to reason in the \prompt~format, we have designed one-shot prompts for different datasets, which are shown Appendix~\ref{app:natural_program_prompt}. Given that LLM's reasoning outputs follow the \prompt~format, we can then verify the deductive validity of a \textit{single} reasoning step $s_i$ through an instruction that consists of (1) the full descriptions of premises used for the reasoning of $s_i$; (2) the full description of $s_i$; (3) an instruction for validity verification, such as ``\texttt{Double-check the reasoning process, let's analyze its correctness, and
end with "yes" or "no".}'' Note that throughout this verification process, we only retain the minimal necessary premise and context for $s_i$, thereby avoiding irrelevant context distraction and significantly improving the effectiveness of validation. Additionally, we employ a one-shot prompt for this verification process, which we find very helpful for improving the verification accuracy. The prompt is shown in Appendix~\ref{app:prompt verification}.

Figure~\ref{fig:natural_program} provides an overview of the complete \prompt-based deductive reasoning and verification process. By using the \prompt~approach, we demonstrate that LLMs are capable of performing explicit, rigorous, and coherent deductive reasoning. Furthermore, \prompt ~enables LLMs to self-verify their reasoning processes more effectively, enhancing the reliability and trustworthiness of the generated responses.

\subsection{Integrating Deductive Verification with Unanimity-Plurality Voting}
\label{sec:method_verification_consistency}
Given that we can \textit{effectively} verify a deductive reasoning process, we can naturally integrate verification with LLM's sequence generation strategies to enhance the trustworthiness of both the intermediate reasoning steps and the final answers. In this work, we propose Unanimity-Plurality Voting, a 2-phase sequence generation strategy described as follows. Firstly, similar to prior work like~\cite{wang2022self}, we sample $k$ reasoning chain candidates along with their final answers. In the unanimity phase, we perform deductive validation on each reasoning chain. Recall that a chain $S$ is valid (i.e., $V(S)=1$) if and only if all of its intermediate reasoning steps are valid (i.e., $\forall i, V(s_i)=1$). For \textit{each} intermediate reasoning step $s_i$, we perform majority voting over $k'$ sampled single-step validity predictions to determine its final validity $V(s_i)$. We then only retain the verified chain candidates $\{S: V(S)=1\}$. In the plurality voting stage, we conduct a majority-based voting among the verified chain candidates to determine the final answer. This voting process ensures that the final answer is selected based on a consensus among the trustworthy reasoning chains.

\section{Experiments}
In this section, we perform evaluations to demonstrate the effectiveness of our Natural Program-based deductive reasoning verification approach over diverse reasoning datasets. Firstly, we show that our deductive verification process leads to substantial improvements in the rigor and reliability of reasoning chains. Subsequently, we will examine the impact of deductive verification on the accuracy of final answers. Our findings reveal that by adopting our Natural Program reasoning format without verification, we improve answer correctness on challenging benchmarks. Further applying deductive verification leads to slight reductions in final answer accuracy. One reason for this phenomenon is that the verification process effectively identifies and eliminates flawed reasoning chains that still produce correct answers.

\subsection{Experimental Setup}

\textbf{Benchmarks.} \label{benchmarks} We evaluate the deductive verification accuracy and the answer correctness of reasoning chains over a diverse set of reasoning tasks: arithmetic reasoning, symbol manipulation, and date understanding. For arithmetic reasoning, we utilize the following benchmarks: 1)~AddSub~\cite{hosseini2014learning}; 2)~GSM8K~\cite{cobbe2021training}; 3)~MATH~\cite{hendrycks2021measuring};  4)~AQuA~\cite{ling2017program}. Among these benchmarks, the AddSub and GSM8K datasets involve middle school-level multi-step calculations to arrive at a single number as the final answer. The MATH dataset presents more challenging problems that require expressing the answer as a mathematical expression in LaTeX format. These problems involve concepts from linear algebra, algebra, geometry, calculus, statistics, and number theory. AQuA also features similarly challenging problems, except that questions are in a multiple-choice format. For symbol manipulation, we use Last Letter Concatenation~\cite{wei2022chain}, where the model is tasked with concatenate the last letters of all the words provided in the question. For date understanding, we use the one from BIG-bench~\cite{srivastava2022beyond}

\textbf{Deductive verfication evaluation setup.} For each of the above benchmarks, we select 100 reasoning chains, where 50 of them are deductively valid and 50 of them exhibit reasoning mistakes. The ground-truth deductive validity of each reasoning chain is determined by human annotators.

\textbf{Answer extraction.} To extract answers from reasoning solutions, we first perform text splitting based on answer prefix patterns such as ``answer is'' or ``option is''. Then, using problem type-specific regular expressions, we extract the final answer. To extract the validity results from deductive verification processes, we only keep the last sentence of model response. We then extract the validity answer with regular expressions to obtain attitude words, e.g., ``yes'' or ``no'', to determine the validity answer. Sometimes, language models may not provide a direct answer and instead output phrases like ``not applicable'' at the end of the response. In such cases, we consider the answer from the model as "yes". Please refer to Appendix~\ref{app:answer_extraction} for more details. 

\textbf{Model and Hyperparameters.} We conduct our main experiments with GPT-3.5-turbo~(ChatGPT)~\cite{ouyang2022training}. We also present results for the LLama model-family~\cite{touvron2023llama}) in Appendix~\ref{app:vicuna_deductive_verification}, where we find the deductive verification accuracy to be worse than larger models even after finetuning. For ChatGPT, we use a generation temperature of $T=0.7$. For Unanimity-Plurality Voting, we set $k=10$ and $k'=3$ by default. We use 1-shot prompting for both reasoning chain generation and deductive verification (except reasoning chain generation for the date understanding task where we use 2-shot). See Appendix~\ref{app:natural_program_prompt} and Appendix~\ref{app:prompt verification} for more details.

\begin{table}[t]
\small
\setlength{\tabcolsep}{2.3pt}
\centering
\begin{tabular}{c|ccccccc|c} 
\toprule
Verification Method                                                                                & Reasoning Correctness   & GSM8k         & AQuA          & MATH          & AddSub        & Date          & Last Letters  & Overall        \\ 
\midrule
\multirow{3}{*}{\begin{tabular}[c]{@{}c@{}}CoT \\ Two-shot \end{tabular}} & Correct & 98\% & 96\% & 100\% & 92\% & 100\% & 96\% & 97\% \\
& Incorrect & 2\% & 4\% & 0\% & 6\% & 26\% & 6\% & 7\% \\
& (Average) & 50\% & 50\% & 50\% & 49\% & 63\% & 51\% & 52\% \\
\midrule
\multirow{3}{*}{\begin{tabular}[c]{@{}c@{}}Natural Program \\
One-shot\end{tabular}}    & Correct & 84\% & 72\%          & 70\%          & 95\%          & 90\% & 96\% & 85\%  \\
& Incorrect   & 84\% & 62\% & 76\% & 40\% & 56\% & 6\%           & 54\%  \\
& (Average) & \textbf{84\%} & \textbf{67\%} & \textbf{73\%} & \textbf{68\%} & \textbf{73\%} & 51\% & \textbf{69\%}  \\ 
\bottomrule
\end{tabular}
\vspace{1em}
\caption{Comparison of deductive verification accuracy of reasoning chains for GPT-3.5-turbo~(ChatGPT). We compare two approaches: (1) verifying entire reasoning chains generated by Chain-of-Thought prompting; (2) verifying reasoning chains generated in the Natural Program format with step-by-step decomposition. In the latter case, when we verify each reasoning step $s_i$, we only keep the necessary subset of premises $\bar{p_i} \subseteq p_i$. To calculate verification accuracy, for each
dataset, we randomly sample 50 reasoning chains that are deductively valid and 50 reasoning steps
exhibiting incorrect reasonings.
}
\label{tab:comparison_verify_methods}
\vspace{-1.5em}
\end{table}

\begin{table}[t]
\setlength{\tabcolsep}{2.6pt}
\centering
\small
\begin{tabular}{ccccccc} 
\toprule
                & \multicolumn{4}{c}{Arithmetic}         & \multicolumn{2}{c}{Commonsense}  \\ 
\cmidrule(lr{1pt}){2-5}\cmidrule(lr{1pt}){6-7}
Methods         & GSM8K   & AQuA    & MATH$^*$ & AddSub  & Date    & Last Letters           \\ 
\midrule
CoT + Voting            & \textbf{87.62\%} & 70.18\% & 35.93\%  & 92.36\% & 69.97\% & 81.60\%                \\
Faithful CoT + Voting   & 75.80\% & 61.80\% &    31.78\%\footnote[1]{\label{results} Most results for Faithful CoT are from their official repository \url{https://github.com/veronica320/Faithful-COT}, except MATH and AddSub due to the unavailability. For these two datasets, we use our implementation and the same prompt for the math word problems in their paper. The prompt for Last Letters is not available, so we leave it blank.} & 88.35\%\footref{results}  &    \textbf{73.50\%}     & -                      \\ 
\midrule
Ours (Natural Program (NP), No Verification) & 87.05\% & \textbf{70.34\%} & \textbf{36.75\%}  & \textbf{93.67\%} & 72.49\% & \textbf{92.98\%}                \\
Ours (NP + Deductive Verification + UPV) & 86.01\% & 69.49\% & 36.48\%  & 93.54\% & 71.45\% & 92.60\%                \\
\bottomrule
\end{tabular}
\vspace{1em}
\caption{Final answer accuracy comparison on GPT-3.5-turbo~(ChatGPT). All approaches generate $k=10$ reasoning chains for each problem before performing majority voting or reasoning chain filtering with our deductive verification approach.}
\label{tab:benchmark_results}
\vspace{-1em}
\end{table}
\subsection{Comparison of Deductive Verification Accuracy}
We compare the verification accuracy of reasoning chains using two methods: (1) verifying the entire reasoning chain at once (as described in Section~\ref{sec:method_decomposition}) without utilizing the Natural Program, and (2) our Natural Program-based verification approach with step-by-step decomposition. The results, presented in Table~\ref{tab:comparison_verify_methods}, indicate that our approach achieves significantly higher reasoning verification accuracy across most datasets. It effectively identifies erroneous reasoning in faulty chains while maintaining a low rate of false positives for valid chains. However, we observe that our approach's effectiveness is limited on the ``Last Letters'' task. We hypothesize that this is due to the task's nature, where each subsequent reasoning step is conditioned on \textit{all} previous steps, presenting greater challenges for reasoning verification due to the increased dependency among premises.

\begin{table*}[t]
\centering
\small
\begin{tabular}{l} 
\toprule
\begin{tabular}[c]{@{}l@{}}\textbf{\textbf{Question}}: In March it rained 0.81 inches. It rained 0.35 inches less in April than in March.\\ How much did it rain in April?\\\textbf{\textbf{Ground Truth:}}~It rained 0.81 - 0.35 = 0.46 inches.\end{tabular} \\ 
\midrule
\textbf{GPT-3.5-turbo~(ChatGPT):} \\
\textbf{Question-related premises:} \\
\#1. It rained 0.81 inches in March. \\
\#2. It rained 0.35 inches less in April~than in March. \\
\#3. How much did it rain in April? \\
\textbf{The first reasoning step:} \\
\#4. \red{(by \#2)} Step 1: Calculate how much it rained in April by subtracting the difference of 0.35 inches from \\the amount in March. \\
Amount of rain in April:~~0.81 - 0.35 = 0.46 inches \\
\bottomrule
\end{tabular}
\caption{For this reasoning chain candidate of this GSM8k problem, ChatGPT generates the correct final answer but provides incorrect premise numbers for grounding the first reasoning step. In ``Step 1'', the correct premise numbers should be \#1 and \#2. Our deductive reasoning verification approach effectively identifies these reasoning errors, enhancing the rigor and trustworthiness of the reasoning process. It is worth noting that removing a correct answer from the candidate reasoning chains has a slightly negative impact on the overall final answer correctness.}
\label{tab:success_case_1}
\end{table*}

\subsection{Impact of Natural Program and Deductive Verification on Final Answer Correctness}
We then investigate the impact of our \prompt~reasoning format and our deductive verification process on final answer correctness. We conduct two experiments: (1) for each problem, we instruct language models to generate $k=10$ reasoning chain candidates in the Natural Program (NP) format and perform simple majority voting on final answers, \textit{without} using deductive verification to filter out reasoning chain candidates; (2) applying our deductive verification approach to filter out reasoning chain candidates, and apply Unanimity-Plurality Voting (UPV) along the process to determine the final answer. As a reference, we also report the performance of Chain-of-Thought (CoT)~\cite{wei2022chain} and Faithful CoT~\cite{lyu2023faithful}. For these baselines, we perform simple answer-based majority voting with $k=10$ for fair comparison. 

Results are presented in Tab.~\ref{tab:benchmark_results}. While our major goal is to improve the trustworthiness and reliability of deductive reasoning, we find that prompting language models to reason in our Natural Program format achieves on-par or better final answer accuracy than baselines over many reasoning tasks. Upon further applying our deductive verification approach to filter out invalid reasoning chains, we observe a slight decrease in final answer accuracy. One major contributing factor to this decrease is the filtering out of reasoning chain candidates that provide correct answers but exhibit incorrect reasoning. We illustrate an example in Table~\ref{tab:success_case_1}, where ChatGPT generates the correct final answer but assigns incorrect premise numbers to support the first reasoning step. We note that in many such cases, our approach effectively identifies these reasoning errors, thereby enhancing the rigor and reliability of the language models' reasoning processes, albeit with a slight negative impact on the overall final answer correctness. Further discussions are presented in  Appendix~\ref{app:more_discussion_verification_answer_correctness}.

\begin{table*}[t]
\centering
\setlength{\tabcolsep}{4.5pt}
\scriptsize
\begin{tabular}{cc|ccccccc|c} 
\toprule
 Premise Context & \# Shots & Reasoning Correctness & GSM8K & AQuA & MATH & AddSub & Date & Last Letters & Average\\ 
\midrule
\multirow{3}{*}{Full Premises} & \multirow{3}{*}{1} & Correct & 64\% & 54\% & 58\% & 95\% & 26\% & 96\% & 66\% \\
 & & Wrong & 56\% & 68\% & 56\% & 24\% & 76\% & 5\% & 48\% \\
 & & (Average) & 60\% & 61\% & 57\% & 60\% & 51\% & 51\% & 57\% \\
 \midrule
\multirow{3}{*}{Minimal Premises} & \multirow{3}{*}{0} & Correct & 84\% & 78\% & 90\% & 96\% & 90\% & 12\%          & 75\% \\
 & & Wrong & 26\%          & 12\%          & 28\%          & 20\%          & 20\%          & 80\% & 31\% \\
 & & (Average) & 55\%          & 45\%          & 59\%          & 58\%          & 55\%          & 46\%          & 53\% \\
 \midrule
\multirow{3}{*}{Minimal Premises} & \multirow{3}{*}{1} & Correct & 84\% & 72\% & 70\% & 95\% & 90\% & 96\% & 85\% \\
 & & Wrong & 84\% & 62\% & 76\% & 40\% & 56\% & 6\% & 54\% \\
 & & (Average) & \textbf{84\%} & \textbf{67\%} & \textbf{73\%} & \textbf{68\%} & \textbf{73\%} & 51\% & \textbf{69\%}  \\ 
\bottomrule
\end{tabular}
\caption{Ablation study on the impact of (1) premise context and (2) zero-shot vs. few-shot scenarios on deductive verification accuracy using our Natural Program-based approach with step-by-step reasoning chain decomposition. To verify each reasoning step $s_i$, we either the full premises $p_i=QC\cup S_{\le j}$, or use the minimal subset of premises $\bar{p_i} \subseteq p_i$ necessary as outlined in Sec.~\ref{sec:method_decomposition} The one-shot prompt we used is shown in Appendix~\ref{app:prompt verification}. For each dataset, we randomly sample 50 reasoning chains that are deductively valid and 50 reasoning steps exhibiting incorrect reasonings. }
\label{tab:multi_step_verification}
\vspace{-1em}
\end{table*}

\subsection{Ablation Study} In addition, we perform several ablation studies to gain further insights into the designs of our deductive verification approach. In Tab.~\ref{tab:multi_step_verification}, we compare two different approaches to verify a single reasoning step $s_i\in S$ following our \prompt~format. The first approach utilizes all premises $p_i=QC\cup S_{\le j}$ for verification regardless of their relevance to $s_i$, potentially introducing irrelevant contexts. The second approach follows our design in Sec.~\ref{sec:method_decomposition} and only includes the necessary context and premises $\bar{p_i} \subseteq p_i$. We observe that removing irrelevant premises significantly improves the reasoning chain verification accuracy on many datasets, highlighting the importance of this technique.

We also ablate on our Unanimity-Plurality Voting strategy by investigating the impact of different $k'$. Recall that $k'$ determines the number of votes to produce validity predictions of single-step reasoning. Results are shown in Tab.~\ref{tab:full_solution}. We observe that increasing $k'$ generally enhances reasoning validation accuracy, though we note that this is at the expense of more compute.

\begin{table*}[t]
\setlength{\tabcolsep}{3.3pt}
\centering
\begin{tabular}{ccccc} 
\toprule
Answer Correctness & $k'=1$ & $k'=3$ & $k'=5$ & $k'=10$ \\\midrule
Correct & 86\% & 90\% & 90\% & 92\% \\\midrule
Wrong & 38\% & 38\% & 38\% & 40\% \\
\bottomrule
\end{tabular}
\caption{Ablation of different values of $k'$ on the verification accuracy of reasoning chains using our Unanimity-Plurality Voting strategy. Experiments are performed on AddSub using GPT-3.5-turbo (ChatGPT).}
\label{tab:full_solution}
\end{table*}

\section{Limitations}
While we have demonstrated the effectiveness of \prompt-based deductive reasoning verification to enhance the trustworthiness and interpretability of reasoning steps and final answers, it is important to acknowledge that our approach has limitations. In this section, we analyze a common source of failure cases to gain deeper insights into the behaviors of our approach. The failure case, as shown in Tab.~\ref{tab:failure_case_1}, involves the ambiguous interpretation of the term ``pennies,'' which can be understood as either a type of coin or a unit of currency depending on the context. The ground truth answer interprets ``pennies'' as coins, while ChatGPT interprets it as a unit of currency. In this case, our deductive verification process is incapable of finding such misinterpretations. Contextual ambiguities like this are common in real-world scenarios, highlighting the current limitation of our approach. 

\begin{table}[t]
\centering
\begin{tabular}{l} 
\toprule
\textbf{Question}: Melanie had 10 quarters and 17 pennies in her bank. Her dad gave her 27 pennies \\
and her mother gave her 19 pennies. How many pennies does Melanie have now? \\
\textbf{Ground Truth}: Melanie have 17 + 27 + 19 = 63 pennies. \\ 
\midrule
\textbf{ChatGPT's reasoning step:} \\
\#5. (by \#1) Step 1: Calculate the number of pennies Melanie had initially. \\
\red{Number of pennies in 10 quarters: 10 * 25 = 250} \\
Number of pennies initially: 250 + 17 = 267 \\
\bottomrule
\end{tabular}
\vspace{1em}
\caption{An example question with ambiguous wordings. The term "pennies" in this question can be interpreted as either a type of coin or a unit of currency. In this particular question, "pennies" is treated as a type of coin. However, the initial reasoning step by ChatGPT mistakenly treats "pennies" as a unit of currency, resulting in the conversion of all Melanie's money into "pennies" (highlighted in red). Consequently, all subsequent reasoning steps follow this flawed logic, leading to an incorrect reasoning trace. Our deductive verification is not yet able to detect such errors.}
\label{tab:failure_case_1}
\vspace{-1em}
\end{table}

\section{Conclusion}
In this paper, we aim to enable Large Language Models (LLMs) to perform explicit and rigorous deductive reasoning while ensuring the trustworthiness of their reasoning processes through self-verification. To this end, we have proposed a novel framework based on ``Natural Program'', a natural language-based deductive reasoning format that facilitates reasoning verification and can be easily generated through in-context learning. Within this framework, we decompose the verification process of complex reasoning chains into step-by-step subprocesses that focus solely on necessary context and premises, allowing us to significantly enhance the accuracy of verification. Additionally, we introduce a Unanimity-Plurality Voting strategy to further improve verification accuracy.  Experimentally, we demonstrate the superiority of our framework in improving the rigor, trustworthiness, and interpretability of reasoning steps and answers. 

\textbf{Broader Impact.} While our deductive verification approach can mitigate hallucinations and reasoning errors of Large Language Models (LLMs), it does not completely eliminate these phenomena. LLMs can still produce harmful and biased content, make incorrect claims, and produce wrongful advice. This issue becomes particularly significant when LLMs engage in complex reasoning chains, increasing the risk of misleading users. Consequently, it is still crucial for users to exercise great caution when interacting with, deploying, or developing LLM-based applications.

\section*{Acknowledgements}
We would like to express our sincere gratitude to Tongzhou Mu and Caiwei Xiao from UC San Diego, Kairong Luo from Tsinghua University, and Pulkit Madan, Reza Pourreza, Sunny Panchal, and Apratim Bhattacharyya from Qualcomm for their valuable discussions and feedback.

{\small
\bibliographystyle{plain}
\bibliography{references}

\begin{thebibliography}{10}

\bibitem{arora2022exposure}
Kushal Arora, Layla~El Asri, Hareesh Bahuleyan, and Jackie Chi~Kit Cheung.
\newblock Why exposure bias matters: An imitation learning perspective of error
  accumulation in language generation.
\newblock {\em arXiv preprint arXiv:2204.01171}, 2022.

\bibitem{bostrom2022natural}
Kaj Bostrom, Zayne Sprague, Swarat Chaudhuri, and Greg Durrett.
\newblock Natural language deduction through search over statement
  compositions.
\newblock {\em arXiv preprint arXiv:2201.06028}, 2022.

\bibitem{brown2020language}
Tom Brown, Benjamin Mann, Nick Ryder, Melanie Subbiah, Jared~D Kaplan, Prafulla
  Dhariwal, Arvind Neelakantan, Pranav Shyam, Girish Sastry, Amanda Askell,
  et~al.
\newblock Language models are few-shot learners.
\newblock {\em Advances in neural information processing systems},
  33:1877--1901, 2020.

\bibitem{bubeck2023sparks}
S{\'e}bastien Bubeck, Varun Chandrasekaran, Ronen Eldan, Johannes Gehrke, Eric
  Horvitz, Ece Kamar, Peter Lee, Yin~Tat Lee, Yuanzhi Li, Scott Lundberg,
  et~al.
\newblock Sparks of artificial general intelligence: Early experiments with
  gpt-4.
\newblock {\em arXiv preprint arXiv:2303.12712}, 2023.

\bibitem{chen2022program}
Wenhu Chen, Xueguang Ma, Xinyi Wang, and William~W Cohen.
\newblock Program of thoughts prompting: Disentangling computation from
  reasoning for numerical reasoning tasks.
\newblock {\em arXiv preprint arXiv:2211.12588}, 2022.

\bibitem{chen2023teaching}
Xinyun Chen, Maxwell Lin, Nathanael Sch{\"a}rli, and Denny Zhou.
\newblock Teaching large language models to self-debug.
\newblock {\em arXiv preprint arXiv:2304.05128}, 2023.

\bibitem{vicuna2023}
Wei-Lin Chiang, Zhuohan Li, Zi~Lin, Ying Sheng, Zhanghao Wu, Hao Zhang, Lianmin
  Zheng, Siyuan Zhuang, Yonghao Zhuang, Joseph~E. Gonzalez, Ion Stoica, and
  Eric~P. Xing.
\newblock Vicuna: An open-source chatbot impressing gpt-4 with 90\%* chatgpt
  quality, March 2023.

\bibitem{chowdhery2022palm}
Aakanksha Chowdhery, Sharan Narang, Jacob Devlin, Maarten Bosma, Gaurav Mishra,
  Adam Roberts, Paul Barham, Hyung~Won Chung, Charles Sutton, Sebastian
  Gehrmann, et~al.
\newblock Palm: Scaling language modeling with pathways.
\newblock {\em arXiv preprint arXiv:2204.02311}, 2022.

\bibitem{chung2022scaling}
Hyung~Won Chung, Le~Hou, Shayne Longpre, Barret Zoph, Yi~Tay, William Fedus,
  Eric Li, Xuezhi Wang, Mostafa Dehghani, Siddhartha Brahma, et~al.
\newblock Scaling instruction-finetuned language models.
\newblock {\em arXiv preprint arXiv:2210.11416}, 2022.

\bibitem{cobbe2021training}
Karl Cobbe, Vineet Kosaraju, Mohammad Bavarian, Mark Chen, Heewoo Jun, Lukasz
  Kaiser, Matthias Plappert, Jerry Tworek, Jacob Hilton, Reiichiro Nakano,
  et~al.
\newblock Training verifiers to solve math word problems.
\newblock {\em arXiv preprint arXiv:2110.14168}, 2021.

\bibitem{creswell2022faithful}
Antonia Creswell and Murray Shanahan.
\newblock Faithful reasoning using large language models.
\newblock {\em arXiv preprint arXiv:2208.14271}, 2022.

\bibitem{creswell2023selectioninference}
Antonia Creswell, Murray Shanahan, and Irina Higgins.
\newblock Selection-inference: Exploiting large language models for
  interpretable logical reasoning.
\newblock In {\em The Eleventh International Conference on Learning
  Representations}, 2023.

\bibitem{driess2023palm}
Danny Driess, Fei Xia, Mehdi~SM Sajjadi, Corey Lynch, Aakanksha Chowdhery,
  Brian Ichter, Ayzaan Wahid, Jonathan Tompson, Quan Vuong, Tianhe Yu, et~al.
\newblock Palm-e: An embodied multimodal language model.
\newblock {\em arXiv preprint arXiv:2303.03378}, 2023.

\bibitem{drozdov2022compositional}
Andrew Drozdov, Nathanael Sch{\"a}rli, Ekin Aky{\"u}rek, Nathan Scales, Xinying
  Song, Xinyun Chen, Olivier Bousquet, and Denny Zhou.
\newblock Compositional semantic parsing with large language models.
\newblock {\em arXiv preprint arXiv:2209.15003}, 2022.

\bibitem{golovneva2022roscoe}
Olga Golovneva, Moya~Peng Chen, Spencer Poff, Martin Corredor, Luke
  Zettlemoyer, Maryam Fazel-Zarandi, and Asli Celikyilmaz.
\newblock Roscoe: A suite of metrics for scoring step-by-step reasoning.
\newblock In {\em The Eleventh International Conference on Learning
  Representations}, 2022.

\bibitem{guerreiro2023hallucinations}
Nuno~M Guerreiro, Duarte Alves, Jonas Waldendorf, Barry Haddow, Alexandra
  Birch, Pierre Colombo, and Andr{\'e}~FT Martins.
\newblock Hallucinations in large multilingual translation models.
\newblock {\em arXiv preprint arXiv:2303.16104}, 2023.

\bibitem{hendrycks2021measuring}
Dan Hendrycks, Collin Burns, Saurav Kadavath, Akul Arora, Steven Basart, Eric
  Tang, Dawn Song, and Jacob Steinhardt.
\newblock Measuring mathematical problem solving with the math dataset.
\newblock {\em arXiv preprint arXiv:2103.03874}, 2021.

\bibitem{hoffmann2022training}
Jordan Hoffmann, Sebastian Borgeaud, Arthur Mensch, Elena Buchatskaya, Trevor
  Cai, Eliza Rutherford, Diego de~Las Casas, Lisa~Anne Hendricks, Johannes
  Welbl, Aidan Clark, et~al.
\newblock Training compute-optimal large language models.
\newblock {\em arXiv preprint arXiv:2203.15556}, 2022.

\bibitem{hosseini2014learning}
Mohammad~Javad Hosseini, Hannaneh Hajishirzi, Oren Etzioni, and Nate Kushman.
\newblock Learning to solve arithmetic word problems with verb categorization.
\newblock In {\em EMNLP}, pages 523--533, 2014.

\bibitem{ji2023survey}
Ziwei Ji, Nayeon Lee, Rita Frieske, Tiezheng Yu, Dan Su, Yan Xu, Etsuko Ishii,
  Ye~Jin Bang, Andrea Madotto, and Pascale Fung.
\newblock Survey of hallucination in natural language generation.
\newblock {\em ACM Computing Surveys}, 55(12):1--38, 2023.

\bibitem{kojima2022large}
Takeshi Kojima, Shixiang~Shane Gu, Machel Reid, Yutaka Matsuo, and Yusuke
  Iwasawa.
\newblock Large language models are zero-shot reasoners.
\newblock {\em arXiv preprint arXiv:2205.11916}, 2022.

\bibitem{kushman-etal-2014-learning}
Nate Kushman, Yoav Artzi, Luke Zettlemoyer, and Regina Barzilay.
\newblock Learning to automatically solve algebra word problems.
\newblock In {\em Proceedings of the 52nd Annual Meeting of the Association for
  Computational Linguistics (Volume 1: Long Papers)}, pages 271--281,
  Baltimore, Maryland, June 2014. Association for Computational Linguistics.

\bibitem{lampinen-etal-2022-language}
Andrew Lampinen, Ishita Dasgupta, Stephanie Chan, Kory Mathewson, Mh~Tessler,
  Antonia Creswell, James McClelland, Jane Wang, and Felix Hill.
\newblock Can language models learn from explanations in context?
\newblock In {\em Findings of the Association for Computational Linguistics:
  EMNLP 2022}, pages 537--563, Abu Dhabi, United Arab Emirates, December 2022.
  Association for Computational Linguistics.

\bibitem{ling2017program}
Wang Ling, Dani Yogatama, Chris Dyer, and Phil Blunsom.
\newblock Program induction by rationale generation: Learning to solve and
  explain algebraic word problems.
\newblock {\em arXiv preprint arXiv:1705.04146}, 2017.

\bibitem{liu2023pre}
Pengfei Liu, Weizhe Yuan, Jinlan Fu, Zhengbao Jiang, Hiroaki Hayashi, and
  Graham Neubig.
\newblock Pre-train, prompt, and predict: A systematic survey of prompting
  methods in natural language processing.
\newblock {\em ACM Computing Surveys}, 55(9):1--35, 2023.

\bibitem{lu2022learn}
Pan Lu, Swaroop Mishra, Tanglin Xia, Liang Qiu, Kai-Wei Chang, Song-Chun Zhu,
  Oyvind Tafjord, Peter Clark, and Ashwin Kalyan.
\newblock Learn to explain: Multimodal reasoning via thought chains for science
  question answering.
\newblock {\em Advances in Neural Information Processing Systems},
  35:2507--2521, 2022.

\bibitem{lyu2023faithful}
Qing Lyu, Shreya Havaldar, Adam Stein, Li~Zhang, Delip Rao, Eric Wong, Marianna
  Apidianaki, and Chris Callison-Burch.
\newblock Faithful chain-of-thought reasoning.
\newblock {\em arXiv preprint arXiv:2301.13379}, 2023.

\bibitem{madaan2023self}
Aman Madaan, Niket Tandon, Prakhar Gupta, Skyler Hallinan, Luyu Gao, Sarah
  Wiegreffe, Uri Alon, Nouha Dziri, Shrimai Prabhumoye, Yiming Yang, et~al.
\newblock Self-refine: Iterative refinement with self-feedback.
\newblock {\em arXiv preprint arXiv:2303.17651}, 2023.

\bibitem{marasovic2022fewshot}
Ana Marasović, Iz~Beltagy, Doug Downey, and Matthew~E. Peters.
\newblock Few-shot self-rationalization with natural language prompts, 2022.

\bibitem{maynez2020faithfulness}
Joshua Maynez, Shashi Narayan, Bernd Bohnet, and Ryan McDonald.
\newblock On faithfulness and factuality in abstractive summarization.
\newblock {\em arXiv preprint arXiv:2005.00661}, 2020.

\bibitem{openai2023gpt4}
OpenAI.
\newblock Gpt-4 technical report, 2023.

\bibitem{ouyang2022training}
Long Ouyang, Jeffrey Wu, Xu~Jiang, Diogo Almeida, Carroll Wainwright, Pamela
  Mishkin, Chong Zhang, Sandhini Agarwal, Katarina Slama, Alex Ray, et~al.
\newblock Training language models to follow instructions with human feedback.
\newblock {\em Advances in Neural Information Processing Systems},
  35:27730--27744, 2022.

\bibitem{paul2023refiner}
Debjit Paul, Mete Ismayilzada, Maxime Peyrard, Beatriz Borges, Antoine
  Bosselut, Robert West, and Boi Faltings.
\newblock Refiner: Reasoning feedback on intermediate representations.
\newblock {\em arXiv preprint arXiv:2304.01904}, 2023.

\bibitem{Prasad2023ReCEval}
Archiki Prasad, Swarnadeep Saha, Xiang Zhou, and Mohit Bansal.
\newblock Receval: Evaluating reasoning chains via correctness and
  informativeness.
\newblock 2023.

\bibitem{ribeiro2023street}
Danilo Ribeiro, Shen Wang, Xiaofei Ma, Henry Zhu, Rui Dong, Deguang Kong,
  Juliette Burger, Anjelica Ramos, William Wang, Zhiheng Huang, et~al.
\newblock Street: A multi-task structured reasoning and explanation benchmark.
\newblock {\em arXiv preprint arXiv:2302.06729}, 2023.

\bibitem{roy2016solving}
Subhro Roy and Dan Roth.
\newblock Solving general arithmetic word problems.
\newblock {\em arXiv preprint arXiv:1608.01413}, 2016.

\bibitem{sanh2021multitask}
Victor Sanh, Albert Webson, Colin Raffel, Stephen~H Bach, Lintang Sutawika,
  Zaid Alyafeai, Antoine Chaffin, Arnaud Stiegler, Teven~Le Scao, Arun Raja,
  et~al.
\newblock Multitask prompted training enables zero-shot task generalization.
\newblock {\em arXiv preprint arXiv:2110.08207}, 2021.

\bibitem{scao2022bloom}
Teven~Le Scao, Angela Fan, Christopher Akiki, Ellie Pavlick, Suzana Ili{\'c},
  Daniel Hesslow, Roman Castagn{\'e}, Alexandra~Sasha Luccioni, Fran{\c{c}}ois
  Yvon, Matthias Gall{\'e}, et~al.
\newblock Bloom: A 176b-parameter open-access multilingual language model.
\newblock {\em arXiv preprint arXiv:2211.05100}, 2022.

\bibitem{schick2023toolformer}
Timo Schick, Jane Dwivedi-Yu, Roberto Dess{\`\i}, Roberta Raileanu, Maria
  Lomeli, Luke Zettlemoyer, Nicola Cancedda, and Thomas Scialom.
\newblock Toolformer: Language models can teach themselves to use tools.
\newblock {\em arXiv preprint arXiv:2302.04761}, 2023.

\bibitem{shen2021generate}
Jianhao Shen, Yichun Yin, Lin Li, Lifeng Shang, Xin Jiang, Ming Zhang, and Qun
  Liu.
\newblock Generate \& rank: A multi-task framework for math word problems.
\newblock {\em arXiv preprint arXiv:2109.03034}, 2021.

\bibitem{shi2023large}
Freda Shi, Xinyun Chen, Kanishka Misra, Nathan Scales, David Dohan, Ed~Chi,
  Nathanael Sch{\"a}rli, and Denny Zhou.
\newblock Large language models can be easily distracted by irrelevant context.
\newblock {\em arXiv preprint arXiv:2302.00093}, 2023.

\bibitem{shi2022language}
Freda Shi, Mirac Suzgun, Markus Freitag, Xuezhi Wang, Suraj Srivats, Soroush
  Vosoughi, Hyung~Won Chung, Yi~Tay, Sebastian Ruder, Denny Zhou, et~al.
\newblock Language models are multilingual chain-of-thought reasoners.
\newblock {\em arXiv preprint arXiv:2210.03057}, 2022.

\bibitem{shinn2023reflexion}
Noah Shinn, Beck Labash, and Ashwin Gopinath.
\newblock Reflexion: an autonomous agent with dynamic memory and
  self-reflection.
\newblock {\em arXiv preprint arXiv:2303.11366}, 2023.

\bibitem{si2022prompting}
Chenglei Si, Zhe Gan, Zhengyuan Yang, Shuohang Wang, Jianfeng Wang, Jordan
  Boyd-Graber, and Lijuan Wang.
\newblock Prompting gpt-3 to be reliable.
\newblock {\em arXiv preprint arXiv:2210.09150}, 2022.

\bibitem{srivastava2022beyond}
Aarohi Srivastava, Abhinav Rastogi, Abhishek Rao, Abu Awal~Md Shoeb, Abubakar
  Abid, Adam Fisch, Adam~R Brown, Adam Santoro, Aditya Gupta, Adri{\`a}
  Garriga-Alonso, et~al.
\newblock Beyond the imitation game: Quantifying and extrapolating the
  capabilities of language models.
\newblock {\em arXiv preprint arXiv:2206.04615}, 2022.

\bibitem{proofwriter}
Oyvind Tafjord, Bhavana Dalvi, and Peter Clark.
\newblock Proofwriter: Generating implications, proofs, and abductive
  statements over natural language.
\newblock In Chengqing Zong, Fei Xia, Wenjie Li, and Roberto Navigli, editors,
  {\em Findings of the Association for Computational Linguistics: {ACL/IJCNLP}
  2021, Online Event, August 1-6, 2021}, volume {ACL/IJCNLP} 2021 of {\em
  Findings of {ACL}}, pages 3621--3634. Association for Computational
  Linguistics, 2021.

\bibitem{touvron2023llama}
Hugo Touvron, Thibaut Lavril, Gautier Izacard, Xavier Martinet, Marie-Anne
  Lachaux, Timoth{\'e}e Lacroix, Baptiste Rozi{\`e}re, Naman Goyal, Eric
  Hambro, Faisal Azhar, et~al.
\newblock Llama: Open and efficient foundation language models.
\newblock {\em arXiv preprint arXiv:2302.13971}, 2023.

\bibitem{wang2022self}
Xuezhi Wang, Jason Wei, Dale Schuurmans, Quoc Le, Ed~Chi, and Denny Zhou.
\newblock Self-consistency improves chain of thought reasoning in language
  models.
\newblock {\em arXiv preprint arXiv:2203.11171}, 2022.

\bibitem{wei2022emergent}
Jason Wei, Yi~Tay, Rishi Bommasani, Colin Raffel, Barret Zoph, Sebastian
  Borgeaud, Dani Yogatama, Maarten Bosma, Denny Zhou, Donald Metzler, et~al.
\newblock Emergent abilities of large language models.
\newblock {\em arXiv preprint arXiv:2206.07682}, 2022.

\bibitem{wei2022chain}
Jason Wei, Xuezhi Wang, Dale Schuurmans, Maarten Bosma, Ed~Chi, Quoc Le, and
  Denny Zhou.
\newblock Chain of thought prompting elicits reasoning in large language
  models.
\newblock {\em arXiv preprint arXiv:2201.11903}, 2022.

\bibitem{welleck2019neural}
Sean Welleck, Ilia Kulikov, Stephen Roller, Emily Dinan, Kyunghyun Cho, and
  Jason Weston.
\newblock Neural text generation with unlikelihood training.
\newblock {\em arXiv preprint arXiv:1908.04319}, 2019.

\bibitem{weng2022verification}
Yixuan Weng, Minjun Zhu, Shizhu He, Kang Liu, and Jun Zhao.
\newblock Large language models are reasoners with self-verification.
\newblock {\em arXiv preprint arXiv:2212.09561}, 2022.

\bibitem{yang2022nlproofs}
Kaiyu Yang, Jia Deng, and Danqi Chen.
\newblock Generating natural language proofs with verifier-guided search.
\newblock In {\em Conference on Empirical Methods in Natural Language
  Processing (EMNLP)}, 2022.

\bibitem{yao2022react}
Shunyu Yao, Jeffrey Zhao, Dian Yu, Nan Du, Izhak Shafran, Karthik Narasimhan,
  and Yuan Cao.
\newblock React: Synergizing reasoning and acting in language models.
\newblock {\em arXiv preprint arXiv:2210.03629}, 2022.

\bibitem{zelikman2022star}
Eric Zelikman, Jesse Mu, Noah~D Goodman, and Yuhuai~Tony Wu.
\newblock Star: Self-taught reasoner bootstrapping reasoning with reasoning.
\newblock 2022.

\bibitem{zeng2022socratic}
Andy Zeng, Adrian Wong, Stefan Welker, Krzysztof Choromanski, Federico Tombari,
  Aveek Purohit, Michael Ryoo, Vikas Sindhwani, Johnny Lee, Vincent Vanhoucke,
  et~al.
\newblock Socratic models: Composing zero-shot multimodal reasoning with
  language.
\newblock {\em arXiv preprint arXiv:2204.00598}, 2022.

\bibitem{zhang2022opt}
Susan Zhang, Stephen Roller, Naman Goyal, Mikel Artetxe, Moya Chen, Shuohui
  Chen, Christopher Dewan, Mona Diab, Xian Li, Xi~Victoria Lin, et~al.
\newblock Opt: Open pre-trained transformer language models.
\newblock {\em arXiv preprint arXiv:2205.01068}, 2022.

\bibitem{zhang2022automatic}
Zhuosheng Zhang, Aston Zhang, Mu~Li, and Alex Smola.
\newblock Automatic chain of thought prompting in large language models.
\newblock {\em arXiv preprint arXiv:2210.03493}, 2022.

\bibitem{zhou2022least}
Denny Zhou, Nathanael Sch{\"a}rli, Le~Hou, Jason Wei, Nathan Scales, Xuezhi
  Wang, Dale Schuurmans, Olivier Bousquet, Quoc Le, and Ed~Chi.
\newblock Least-to-most prompting enables complex reasoning in large language
  models.
\newblock {\em arXiv preprint arXiv:2205.10625}, 2022.

\bibitem{zhou2022teaching}
Hattie Zhou, Azade Nova, Hugo Larochelle, Aaron Courville, Behnam Neyshabur,
  and Hanie Sedghi.
\newblock Teaching algorithmic reasoning via in-context learning.
\newblock {\em arXiv preprint arXiv:2211.09066}, 2022.

\end{thebibliography}
}

\appendix

\newpage
\section{Deductive Verification with Vicuna Models}
\label{app:vicuna_deductive_verification}

We further explore the efficacy of deductive verification for open-source models. We select two popular models: Vicuna-7B and Vicuna-13B~\citep{vicuna2023}. These models are fine-tuned versions of LLaMA-7B and LLaMA-13B~\citep{touvron2023llama} using the ShareGPT data\footnote{https://github.com/domeccleston/sharegpt}. We use the same Natural Program-based one-shot verification method we used in the main paper. Results are shown in the first and the third rows of Table~\ref{tab:vicuna results}. We observe for the \textit{original Vicuna models without finetuning}, Vicuna-7B exhibits poor performance in deductive verification and fails to find out reasoning mistakes, while the larger Vicuna-13B exhibits better verification accuracy.

\begin{table}[h]
\centering
\small
\setlength{\tabcolsep}{3.2pt}
\begin{tabular}{c|ccccccc|c} 
\toprule
Models & Reasoning Correctness & GSM8K & AQuA & MATH & AddSub & Date & Last Letters & Overall \\ 
\midrule
\multirow{3}{*}{Vicuna-7B} & Correct & 80\% & 86\% & 96\% & 98\% & 96\% & 80\% & 89\% \\
 & Wrong & 14\% & 22\% & 16\% & 6\% & 20\% & 34\% & 19\% \\
 & (Average) & 47\% & 54\% & 56\% & 52\% & 58\% & 57\% & 54\% \\ 
\midrule
\multirow{3}{*}{\begin{tabular}[c]{@{}c@{}}Vicuna-7B\\(fine-tuned)\end{tabular}} & Correct & 68\% & 48\% & 46\% & 76\% & 46\% & 32\% & 53\% \\
 & Wrong & 72\% & 86\% & 54\% & 60\% & 72\% & 68\% & 69\% \\
 & (Average) & 70\% & \textbf{67\%} & 50\% & 68\% & 61\% & 50\% & 61\% \\ 
\midrule
\multirow{3}{*}{Vicuna-13B} & Correct & 86\% & 82\% & 92\% & 96\% & 72\% & 74\% & 84\% \\
 & Wrong & 32\% & 36\% & 20\% & 20\% & 34\% & 30\% & 29\% \\
 & (Average) & 59\% & 59\% & 56\% & 58\% & 53\% & 52\% & 57\% \\ 
\midrule
\multirow{3}{*}{\begin{tabular}[c]{@{}c@{}}Vicuna-13B\\(fine-tuned)\end{tabular}} & Correct & 74\% & 50\% & 56\% & 86\% & 72\% & 12\% & 58\% \\
 & Wrong & 72\% & 76\% & 72\% & 68\% & 62\% & 96\% & 74\% \\
 & (Average) & \textbf{73\%} & 63\% & \textbf{64\%} & \textbf{77\%} & \textbf{67\%} & \textbf{54\%} & \textbf{66\%} \\ 
\midrule
\midrule
\multirow{3}{*}{\begin{tabular}[c]{@{}c@{}}ChatGPT\\(GPT-3.5-Turbo)\end{tabular}} & Correct & 84\% & 72\% & 70\% & 95\% & 90\% & 96\% & 85\% \\
 & Wrong & 84\% & 62\% & 76\% & 40\% & 56\% & 6\% & 54\% \\
 & (Average) & 84\% & 67\% & 73\% & 68\% & 73\% & 51\% & 69\% \\
\bottomrule
\end{tabular}
\vspace{1em}
\caption{One-shot Deductive Verification Accuracy of Vicuna-7B and Vicuna-13B. The models are evaluated with or without finetuning on our deductive verification dataset. For each
dataset, we randomly sample 50 reasoning chains that are deductively valid and 50 reasoning steps
exhibiting incorrect reasonings.}
\label{tab:vicuna results}
\end{table}

\begin{table}[t]
\centering
\small
\begin{tabular}{r|cc}
\toprule
Hyperparameters & Value\\
\midrule
Optimizer & AdamW \\
Learning rate & $1 \times 10^{-5}$ \\
Weight decay & 0.00 \\
Num epochs & 3 \\
Batch size & 64 \\
Learning rate schedule & Linear \\
\bottomrule
\end{tabular}
\vspace{1em}
\caption{Hyperparameters for finetuning Vicuna models with our deductive verification dataset.}
\label{tab:training param}
\end{table}

We therefore conduct an additional experiment to investigate if the verification accuracy of Vicuna models can be improved by fine-tuning. To this end, we generate a deductive verification dataset, which consists of 2000 reasoning steps evenly distributed between correct and incorrect categories. We automatically generate this dataset using GPT-3.5-turbo since it exhibits a very high accuracy of single-step verification. We first use GPT-3.5-turbo to generate solutions for problems in GSM8K's training set. We then execute step-by-step deductive verification on these solutions using GPT-3.5-turbo. For solutions that result in correct final answers, we retain the reasoning steps that pass deductive verification. For solutions that yield incorrect final answers, we retain the reasoning steps that cannot pass deductive verification. After constructing our dataset, we then fine-tune the Vicuna models using the verifications of the 2000 reasoning steps. Models were fine-tuned with 4 A100-80GB over 3 epochs. Training parameters are shown in Table~\ref{tab:training param}.

As shown in Tab.~\ref{tab:vicuna results}, we observe that fine-tuning with our dataset can enhance the deductive verification accuracy of Vicuna models not only on the dataset where the training dataset is constructed (GSM8K), but also on many other datasets. However, the accuracy is still worse than non-finetuned GPT-3.5, which suggests that model capacity has a significant impact on deductive verification capabilities.

\section{More Discussion on Improvements of Deductive Verification Accuracy Versus Improvements on Final Answer Correctness}
\label{app:more_discussion_verification_answer_correctness}

In the main paper, we demonstrated that our verification approach significantly improves the verification accuracy of reasoning chains (Tab.~\ref{tab:comparison_verify_methods},~\ref{tab:multi_step_verification}, but barely improves the final answer accuracy (Tab.~\ref{tab:benchmark_results}). We further analyze this phenomenon below:

Consider the GSM8K dataset as an example (recall that the final answer for a problem is obtained through majority voting). Among all problems, 91.6\% of problems have $|$(number of votes received by the correct answer) $-$ (largest number of votes received by a single wrong answer)$| > 2$, and their final answers are unlikely to be changed through our deductive verification approach. For the rest of the cases (8.4\%), where deductive verification is more likely to impact their final answers, we found that:
\begin{itemize}
\item Among all reasoning chains that arrive at correct answers (these correct-answer chains account for 49.4\% of all reasoning chain candidates), 46.2\% of reasoning chains are filtered out by our verification process.
\item Among the reasoning chains that arrive at correct answer but are filtered out by our verification process, 76.3\% indeed exhibit incorrect reasoning.
\item Among the reasoning chains that arrive at correct answer and are not filtered out by our verification process, 78.0\% indeed have correct reasonings.
\item Among the reasoning chains that do not arrive at correct answer and exhibit incorrect reasonings (these account for 50.6\% of all reasoning chain candidates), 40.6\% are filtered out by our verification process.
\end{itemize}

The above statistics shows that a significant portion of reasoning chains that arrive at correct answers but exhibit incorrect reasoning are successfully eliminated. Therefore, the reliability and trustfulness of reasoning chains that arrive at the correct answers are significantly improved. Combined with the fact that a significant proportion of reasoning chains that exhibit incorrect answers are eliminated, and that our approach's verification accuracy significantly improves over naive verification approaches, our primary goal to improve LLM reasoning reliability is accomplished.

Nevertheless, the removals of many reasoning chains yielding correct answers (specifically, a significant 46.2\% $\times$ 49.4\% of all chains) has a notable impact. This even exceeds the removals of reasoning chains with incorrect reasonings and answers (40.6\% $\times$ 50.6\% of all chains). As a result, there are fewer votes for the correct answer when generating final answers through majority voting, which limits the final answer accuracy. In the future, we believe that when a greater proportion of incorrect reasoning chains with incorrect answers are filtered out, we can improve the final answer accuracy.

\section{More Details on Answer Extraction}
\label{app:answer_extraction}
In this section, we describe our process to extract the final answer from language models' responses. The process begins by selecting the last three non-empty lines. Then, these lines are processed through the following pipeline:

\begin{enumerate}
    \item Firstly, we use a list of regular expressions to identify "No-Answer" patterns within the text, such as "we cannot answer (this|the) question". This process helps us ascertain whether the model can provide a conclusive answer. If any such patterns appear in the text, we mark "No answer!" as the final answer. However, if we don't detect these patterns, we proceed to the next steps for extracting the final answer.
    \item Secondly, if any "Answer-Split" patterns are found in the text, we divide the text into several blocks using the identified pattern. The last block of text is then utilized for extracting the answer.
    \item Lastly, we use regular expressions, as outlined in Tab.~\ref{tab:final_answer_extraction}, to scan the remaining text for possible final answers. If multiple matches are found for the pattern, we select the first match as the final answer. If no pattern matches are found in the remaining text, we default the final response to "No answer!".
\end{enumerate}

\textbf{``No-Answer'' Patterns:} \label{no-answer-pattern} "we cannot provide an answer to this question with (this$\vert$the) given information", "we cannot answer (this$\vert$the) question", "we cannot determine", "we can't determine", "we do not have enough information to answer (this$\vert$the) question", "we do not have enough information to provide a definitive answer to (this$\vert$the) question", "the answer(.*?)is unknown", "answer is not listed among the answer choices".

\textbf{``Answer-Split'' Patterns:} \label{answer-split-pattern} "answer is", "final answer:", "answer to the question is", "answer to this question is", "concatenated letters are", "concatenate the letters -", "The answer of ".

\begin{table}[h]
\centering
\begin{tabular}{c|c} 
\toprule
Answer Type & Regular Expression \\
\midrule
Number & ~(-?$\backslash$d[$\backslash$d,$\backslash$. ]*) \\
Fractional number & $($-?$\backslash$($\backslash$d+$\backslash/\backslash$d+$\backslash$)$\backslash/\backslash$d+$\vert$-?$\backslash$d+$\backslash/\backslash$d+$)$ \\
Date & $(\backslash$d$\backslash$d$\backslash/\backslash$d$\backslash$d$\backslash/\backslash$d$\backslash$d$\backslash$d$\backslash$d$)$ \\
Yes or No & $($?:Yes$\vert$No$\vert$yes$\vert$no$\vert$NO$\vert$YES$)$ \\
\bottomrule
\end{tabular}
\vspace{1em}
\caption{Regular Expression for extracting the final answers of different kinds of questions.}
\label{tab:final_answer_extraction}
\end{table}

\section{Prompts}

\subsection{Prompt for Direct Reasoning Chain Verification Without Natural Program Format}
\label{app:no_natural_program_prompt}
For the results in Tab.~\ref{tab:cot_verification} of the main paper, We use ``Do you think the above reasoning process is correct? Let's think step by step.'' as the zero-shot prompt to verify an entire reasoning chain at once. We also design a two-shot prompt for reasoning chain verification as shown in Tab.~\ref{tab:prompt chain verification}, which covers one correct reasoning chain and one incorrect reasoning chain.

\subsection{Prompts for Reasoning Chain Generation in the Natural Program Format}
\label{app:natural_program_prompt}
To instruct models to generate reasoning chains in the Natural Program format that facilitates step-by-step deductive verification, we have designed four distinct prompts to address different types of problems. These include:
\begin{enumerate}
    \item Math word problems, as illustrated in Tab.~\ref{tab:prompt mwp}, covering GSM8K, MATH, and AddSub datasets.
    \item Math word problems with multiple-choice options, illustrated in Tab.~\ref{tab:prompt mwp mc}, covering the AQuA dataset.
    \item Date-related problems, illustrated in Tab.~\ref{tab:prompt date}, covering the Date dataset.
    \item Last Letters problems, illustrated in Tab.~\ref{tab:prompt last letters}, covering the Last Letters dataset.
\end{enumerate}

\subsection{Prompt for Deductive Verification Following Natural Program Format and Step-by-Step Decomposition}
\label{app:prompt verification}
We have designed a general one-shot prompt for the deductive verification of a single reasoning step on different datasets, as shown in Tab.~\ref{tab:prompt verification}. This prompt serves to instruct language models to generate the deductive validity of each reasoning step as illustrated in Sec.~\ref{sec:method_natural_program} and the top-right box of Fig.~\ref{fig:natural_program} of the main paper.

\section{More Deductive Verification Examples}

In this section, we present more deductive verification examples using our Natural Program-based approach on single reasoning steps. 

In Tab.~\ref{tab:gsm8k false to false verification}, we demonstrate that the language model (ChatGPT) not only successfully identifies ungrounded information, but also identifies logical errors within the given solutions.

In Tab.~\ref{tab:gsm8k false to true verification}, we illustrate a case where the language model fails to detect ungrounded premise numbers, mistakenly assuming that these numbers can be derived from grounded ones.

Lastly, in Tab.~\ref{tab:gsm8k true to false verification}, we illustrate a case where the language model is sometimes unable to correctly identify grounded numbers.

\newpage
\begin{table*}[t]
\centering
\small
\begin{tabular}{l} 
\toprule
\textbf{Here is a math question and its solution:}\\
"Question:\\
Jason had 20 lollipops. He gave Denny some lollipops. Now Jason has 12 lollipops. How many lollipops \\did Jason give to Denny?\\
\\
Answer:\\
Jason started with 20 lollipops. Then he had 12 after giving some to Denny. So he gave Denny 20 + 12 =34.\\ The answer is 34."\\
\\
You are a math teacher. Do you think the reasoning process is correct?\\
Let's check the "Answer" step by step, and then end with "yes" or "no".\\
\\
\textbf{Answer:}\\
The problem statement is clear, but the reasoning process has a major error in it. \\Let's go through it step by step:\\
\\
1. "Jason started with 20 lollipops." This is correct according to the problem statement.\\
\\
2. "Then he had 12 after giving some to Denny." This is also correct as per the problem.\\
\\
3. "So he gave Denny 20 + 12 =34. The answer is 34." This is incorrect. The correct reasoning process \\ should be subtracting the number of lollipops he ended up with from the number he started with.\\ That would be 20 - 12 = 8 lollipops.\\
\\
So, Jason gave Denny 8 lollipops, not 34. No, the reasoning process is incorrect.\\
\\
\textbf{Here is a math question and its solution:}\\
"Question:\\
Olivia has \$23. She bought five bagels for \$3 each. How much money does she have left?\\
\\
Answer:\\
Olivia had 23 dollars. 5 bagels for 3 dollars each will be 5 x 3 = 15 dollars. So she has 23 - 15 dollars left. \\23 - 15 is 8. The answer is 8."\\
\\
You are a math teacher. Do you think the reasoning process is correct?\\
Let's check the "Answer" step by step, and then end with "yes" or "no".\\
\\
\textbf{Answer:}\\
Let's go through this reasoning process step by step.\\
\\
1. "Olivia had 23 dollars." This is the given starting amount of money that Olivia has. This step is correct.\\
\\
2. "5 bagels for 3 dollars each will be 5 x 3 = 15 dollars." This is correct as well. Olivia spends 3 dollars \\for each bagel, and she buys 5 of them. So the total cost of the bagels is 3 dollars/bagel x 5 bagels = 15 dollars.\\
\\
3. "So she has 23 - 15 dollars left." This statement is also correct. After purchasing the bagels, Olivia will\\ have her initial amount of money minus the cost of the bagels.\\
\\
4. "23 - 15 is 8. The answer is 8." This step is correct as well. 23 dollars (Olivia's initial amount) - 15 dollars\\ (the cost of the bagels) = 8 dollars. This means that Olivia has 8 dollars left after buying the bagels.\\
\\
So overall, Olivia has 8 dollars left after buying five bagels for 3 dollars each. Yes, the reasoning process \\is correct.\\
\bottomrule
\end{tabular}
\caption{Two-shot prompt for direct reasoning chain verification without Natural Program format.}
\label{tab:prompt chain verification}
\end{table*}

\begin{table*}[ht]
\centering
\begin{tabular}{l} 
\toprule
\textbf{Question:}\\
Alice buys 5 apples. Each apple costs her \$2. She gets them for 50\% off. After she bought \\she decides she doesn't like 2 of them and sells them to her friends for 1. How much money\\ was she out? \\
\\
\textbf{Answer:}\\
First, let's write down all the statements and relationships in the question with labels.\\
\#1. Alice gets 5 apples.\\
\#2. Each apple costs \$2, but Alice gets apples for 50\% off.\\
\#3. Alice sells 2 of 5 apples for 1.\\
\#4. How much money was Alice out?\\
\\
Next, let's answer the question step by step with reference to the question and reasoning process:\\
\\
\#5. (by \#2) Step 1: Calculate the price per apple after a 50\% discount.\\
Original price per apple: \$2\\
Discount: 50\%\\
Price per apple after discount: \$2 * (1 - 0.50) = \$2 * 0.50 = \$1\\
\\
\#6. (by \#1 \#5) Step 2: Calculate Alice's cost of the 5 apples after a 50\% discount.\\
Price per apple after discount: \$1\\
Total apples: 5\\
Alice's total cost of 5 apples: \$1 * 5 = \$5\\
\\
\#7. (by \#3) Step 3: Calculate the money Alice gets back from selling 2 apples.\\
Money that Alice gets back from selling 2 apples: \$1\\
\\
\#8. (by \#6 \#7) Step 4: Calculate the total amount of money Alice is out.\\
Alice's total cost of 5 apples: \$5\\
Money that Alice gets back from selling 2 apples: \$1\\
Money that Alice is out: \$5 - \$1 = \$4\\
\\
\#9. (by \#4 \#8) The original question is \#4.\\
How much money was Alice out? We do not miss information on the rewritten labels. So the \\answer to this question is Alice is out \$4.\\
\\
Please answer the following question by following the above template. \\
\bottomrule
\end{tabular}
\caption{One-shot Natural Program prompt for reasoning chain \textit{generation} on math word problems.}
\label{tab:prompt mwp}
\end{table*}

\begin{table*}[t]
\centering
\begin{tabular}{l} 
\toprule
\textbf{Question:}\\
Alice buys 5 apples. Each apple costs her \$2. \\She gets them for 50\% off. After she bought She decides she doesn't like 2 of them and sells them \\to her friends for 1. \\How much money was she out?\\
Answer Choices:\\
A) 4\\
B) 2\\
C) None of the above\\
\\
\textbf{Answer:}\\
First, let's write down all the statements and relationships in the question with labels.\\
\#1. Alice gets 5 apples.\\
\#2. Each apple costs \$2, but Alice gets apples for 50\% off.\\
\#3. Alice sells 2 of 5 apples for 1.\\
\#4. How much money was Alice out?\\
\\
Next, let's answer the question step by step with reference to the question and reasoning process:\\
\\
\#5. (by \#2) Step 1: Calculate the price per apple after a 50\% discount.\\
Original price per apple: \$2\\
Discount: 50\%\\
Price per apple after discount: \$2 * (1 - 0.50) = \$2 * 0.50 = \$1\\
\\
\#6. (by \#1 \#5) Step 2: Calculate Alice's cost of the 5 apples after a 50\% discount.\\
Price per apple after discount: \$1\\
Total apples: 5\\
Alice's total cost of 5 apples: \$1 * 5 = \$5\\
\\
\#7. (by \#3) Step 3: Calculate the money Alice gets back from selling 2 apples. \\
Money that Alice gets back from selling 2 apples: \$1\\
\\
\#8. (by \#6 \#7) Step 4: Calculate the total amount of money Alice is out. \\
Alice's total cost of 5 apples: \$5\\
Money that Alice gets back from selling 2 apples: \$1\\
Money that Alice is out: \$5 - \$1 = \$4\\
\\
\#9. (by \#4 \#8) The original question is \#4. How much money was Alice out? We do not miss\\ information on the rewritten labels. So the answer to this question is Alice is out \$4. Among \\all the answer choices, the best option is A) 4.\\
\\
Please answer the following question by following the above template. \\
\bottomrule
\end{tabular}
\caption{One-shot Natural Program prompt for reasoning chain generation on math word problems with multiple choice.}
\label{tab:prompt mwp mc}
\end{table*}

\begin{table*}[t]
\centering
\begin{tabular}{l} 
\toprule
Let's work on date computation.\\
\textbf{Question:}\\
Yesterday is 02/22/2012. What is the date one week from today? Please answer in format \\MM/DD/YYYY.\\
\\
\textbf{Answer:}\\
First, let's write down all the statements and relationships in the question with labels.\\
\#1. Yesterday is 02/22/2012.\\
\#2. What is the date one week from today? Please answer in format MM/DD/YYYY.\\
\\
Next, let's answer the question step by step with reference to the question and reasoning process:\\
\\
\#4. (by \#1) Step 1: Calculate the date today.\\
The date of Yesterday is 02/22/2012.\\
The date of today is 02/23/2012.\\
\\
\#5. (by \#2 \#4) Step 2: Calculate the date one week from today and write the answer in \\MM/DD/YYYY.\\
Because 2012 is a leap year and February in 2012 has 29 days, the date one week from today is \\03/01/2012.\\
Therefore, date one week from today is 03/01/2012.\\
\\
\textbf{Question:}\\
Today is 3/2 and Bob is watching NBA 2007. What is the date one week before today? \\Please answer in format MM/DD/YYYY.\\
\\
\textbf{Answer:}\\
First, let's write down all the statements and relationships in the question with labels.\\
\#1. Today is 3/2.\\
\#2. Bob is watching NBA 2007 today.\\
\#3. What is the date one week before today? Please answer in format MM/DD/YYYY.\\
\\
Next, let's answer the question step by step with reference to the question and reasoning process:\\
\\
\#4. (by \#1 \#2) Step 1: Calculate the date today.\\
Today is 3/2 and Bob is watching NBA 2007 today. Thus, Today is 03/02/2007.\\
\\
\#5. (by \#3 \#4) Step 2: Calculate the date one week before today and write the answer \\in MM/DD/YYYY.\\
Because 2007 is not a leap year and February in 2007 has 28 days, the date one week from today \\is 02/23/2007.\\
Therefore, the date one week before today is 02/23/2007.\\
\\
Please answer the following question by following the above template.\\
\bottomrule
\end{tabular}
\caption{Two-shot Natural Program prompt for reasoning chain generation on the Date dataset.}
\label{tab:prompt date}
\end{table*}

\begin{table*}[t]
\centering
\begin{tabular}{l} 
\toprule
Let's work on date computation.\\
\textbf{Question:}\\
Take the last letters of each words in "Apple Is" and concatenate them.\\
\\
\textbf{Answer:}\\
First, let's write down all the statements and relationships in the question with labels.\\
\\
\#1. We need to take the last letter of the word "Apple".\\
\#2. We need to take the last letter of the word "Is".\\
\#3. We need to concatenate the last letters of the words sequentially\\
\\
Next, let's answer the question step by step with the complete reference to the question and reasoning process:\\
\\
\#4. (by \#1, \#3) Step 1: Taking the last letter of the word "Apple" and concatenate the latter.\\
The last letter of the word "Apple" is "e".\\
The concatenated last letters are "e".\\
\\
\#5. (by \#2, \#3, \#4) Step 2: Taking the last letter of the word "Is" and concatenate the latter.\\
The last letter of the word "Is" is "s".\\
The concatenated last letters are "es".\\
\\
\#6. (by \#3, \#5) Step 3: Provide the final answer of the concated letter.\\
Therefore, the answer is "es".\\
\\
Please answer the following question by following the above template.\\
\bottomrule
\end{tabular}
\vspace{1em}
\caption{One-shot Natural Program prompt for reasoning chain generation on the Last Letters dataset.}
\label{tab:prompt last letters}
\end{table*}

\begin{table*}[t]
\vspace{-5mm}
\centering
\small
\begin{tabular}{l} 
\toprule
\textbf{Here is some information:}\\
``1. Alice gets 2 apples.\\
2. Alice gets twice as many oranges as bananas.\\
3. Original price per apple: \$2\\
Discount: 50\%\\
Price per apple after discount: \$2 * (1 - 0.50) = \$2 * 0.50 = \$1"\\
\\
Based on the given information, here is a reasoning process:\\
"Calculate Alice's cost of the 5 apples after a 50\% discount.\\
Price per apple after discount: \$1\\
Apples be eaten: 3\\
Oranges be eaten: 6\\
Total apples: 5\\
Total oranges: x\\
Total bananas: 2x\\
Alice's total number of fruits: 5 + x + 2x = 5 + 4x\\
Alice's total cost of 5 apples: \$1 * (5 - 3) = \$2''\\
\\
Double-check the reasoning process, let's analyze its correctness, and end with "yes" or "no".\\
\\
\textbf{Answer:}\\
Let's think step by step.\\
\\
\textcolor{orange}{Grounding check}\\
\textcolor{orange}{Purpose: Find references for numbers in the reasoning process sequentially}\\
\textcolor{orange}{Rule: Any ungrounded number makes the check fails}\\
\\
"x" is a variable, we do not care about variables.\\
"y" is a variable, we do not care about variables.\\
\\
"5" is a quantity from the given information that represents the number of apples.\\
"50\%" is a quantity from the given information that represents the discount.\\
"\$1" is a quantity from the given information that represents the price per apple after the discount.\\
"3" is a quantity not from the given information, it is ungrounded.\\
"6" is a quantity not from the given information, it is ungrounded.\\
"\$2" is a quantity from the given information that represents the original price per apple.\\
Result: Based on the check, there are ungrounded numbers, so the grounding check fails.\\
\\
\textcolor{red}{Reasoning check:}\\
\textcolor{red}{Purpose: Check whether the reasoning logic correctly answers the question}\\
\textcolor{red}{Rule: Any logic error makes the check fails}\\
\\
To calculate the total cost of apples after a 50\% discount, we should multiply the number of apples\\by the discounted price. But to answer the question, we do not need to calculate the total number\\of fruit Alice gets.\\
Result: Based on the check, there are some logic mistakes, so the reasoning check fails.\\
\\
\textcolor{violet}{Calculation check:}\\
\textcolor{violet}{Purpose: Check whether the calculation process is consistent}\\
\textcolor{violet}{Rule: Any inconsistent calculation makes the check fails}\\
\\
calculation1: \\
equation: \$1 * (5 - 3), answer: \$2\\
(5 - 3)  = 2\\
\$1 * 2 = \$2 is consistent with the answer, so the calculation is correct.\\
calculation2:\\
equation: 5 + x + 2x, answer: 5 + 4x\\
x + 2x = 3x\\
5 + 3x is inconsistent with the answer, so the calculation is incorrect.\\
Result: Based on the check, the calculation process is inconsistent, so the calculation check fails.\\
\\
Check results: Ground check fails, Reasoning check fails, Calculation check fails.\\
Rule: Any failed check makes the reasoning incorrect.\\
So the answer is "no".\\
\bottomrule
\end{tabular}
\caption{One-shot prompt for deductive verification of a single reasoning step, following our Natural Program format and step-by-step reasoning chain decomposition.}
\label{tab:prompt verification}
\end{table*}

\begin{table*}[t]
\centering
\begin{tabular}{l} 
\toprule
\textbf{Input:}\\
Here is some information:\\
"For every 25 telephone calls he makes, he gets one person to come into the dealership."\\
\\
Based on the given information, here is a reasoning process:\\
"Calculate the number of people that would come into the dealership for 15 car sales.\\
Number of people needed to sell 15 cars: 15 * 2 = 30"\\
\\
Double-check the reasoning process, let's analyze its correctness, and end with "yes" or "no".\\
\\
Answer:\\
Let's think step by step without any assumptions.\\
\midrule
\textbf{Output:}\\
Grounding check:\\
Purpose: Find references for numbers in the reasoning process sequentially\\
Rule: Any ungrounded number makes the check fails\\
\\
"15" is a quantity from the question that represents the number of car sales.\\
\red{"2" is a quantity not from the given information, it is ungrounded.}\\
"25" is a quantity from the given information that represents the number of telephone calls \\per one person to come into the dealership.\\
Result: \red{Based on the check, there are ungrounded numbers, so the grounding check fails.}\\
\\
Reasoning check:\\
Purpose: Check whether the reasoning logic correctly answers the question\\
Rule: Any logic error makes the check fails\\
\\
It is not clear how the number of people coming into the dealership relates to the number of car sales. \\The given information only tells us the ratio of telephone calls to people coming into the dealership,\\ but it does not provide any information about the number of telephone calls made for each car sale. \\Therefore, the reasoning process is not correct.\\
Result: \red{Based on the check, there is a logic mistake, so the reasoning check fails.}\\
\\
Calculation check:\\
Purpose: Check whether the calculation process is consistent\\
Rule: Any inconsistent calculation makes the check fails\\
\\
calculation1: \\
equation: 15 * 2, answer: 30\\
15 * 2 = 30 is consistent with the answer, so the calculation is correct.\\
Result: Based on the check, the calculation process is consistent, but the reasoning is incorrect, \\so the calculation check fails.\\
\\
Check results: Grounding check fails, Reasoning check fails, Calculation check fails.\\
Rule: Any failed check makes the reasoning incorrect.\\
So the answer is "no".\\
\bottomrule
\end{tabular}
\vspace{1em}
\caption{Successful case: our deductive verification approach successfully discovers ungrounded information and reasoning mistakes.}
\label{tab:gsm8k false to false verification}
\end{table*}

\begin{table*}[t]
\centering
\begin{tabular}{l} 
\toprule
\textbf{Input:}\\
Here is some information:\\
"How many hours would it take for 5 people to paint a whole house?"\\
\\
Based on the given information, here is a reasoning process:\\
"Calculate the total number of hours it would take for 5 people to paint a whole house.\\
Total hours for one person to paint a whole house: 240 hours\\
Total number of people: 5\\
Total hours for 5 people to paint a whole house: 240 hours / 5 = 48 hours"\\
\\
Double-check the reasoning process, let's analyze its correctness, and end with "yes" or "no".\\
\\
Answer:\\
Let's think step by step without any assumptions.\\
\midrule
\textbf{Output:}\\
Grounding check\\
Purpose: Find references for numbers in the reasoning process sequentially\\
Rule: Any ungrounded number makes the check fails\\
\\
\red{"240" is a quantity not from the given information, it is ungrounded.}\\
"5" is a quantity from the given information that represents the number of people.\\
"48" is a quantity calculated from the ungrounded number and grounded number in the reasoning process.\\
Result: Based on the check, there is one ungrounded number, \red{but it is calculated from another grounded} \\\red{number, so the grounding check passes.}\\
\\
Reasoning check:\\
Purpose: Check whether the reasoning logic correctly answers the question\\
Rule: Any logic error makes the check fails\\
\\
The reasoning process correctly calculates the total number of hours for 5 people to paint a whole house by \\dividing the total hours for one person to paint a whole house by 5. Therefore, the reasoning check passes.
\\
Calculation check:\\
Purpose: Check whether the calculation process is consistent\\
Rule: Any inconsistent calculation makes the check fails\\
\\
calculation1: \\
equation: 240 / 5, answer: 48\\
240 / 5 = 48 is consistent with the answer, so the calculation is correct.\\
Result: Based on the check, the calculation process is consistent, so the calculation check passes.\\
\\
Check results: Ground check passes, Reasoning check passes, Calculation check passes.\\
Rule: All checks pass make the reasoning correct.\\
So the answer is "yes".\\
\bottomrule
\end{tabular}
\vspace{1em}
\caption{Failure case: our deductive verification process fails to find out ungrounded information in the reasoning step. The number 240 in the reasoning step is ungrounded, but the model states that it can be calculated from grounded numbers.}
\label{tab:gsm8k false to true verification}
\end{table*}

\begin{table*}[t]
\centering
\begin{tabular}{l} 
\toprule
\textbf{Input:}\\
Here is some information:\\
"Calculate the number of doughnuts Derrick makes in a day.\\
Number of doughnuts in a dozen: 12\\
\red{Number of doughnuts in 10 dozens: 10 * 12 = 120}\\
Calculate the revenue from selling one doughnut.\\
Price per doughnut: \$2"\\
\\
Based on the given information, here is a reasoning process:\\
"Calculate the revenue from selling all the doughnuts in a day.\\
Number of doughnuts in a day: 120\\
Revenue from selling all the doughnuts in a day: \$2 * 120 = \$240"\\
\\
Double-check the reasoning process, let's analyze its correctness, and end with "yes" or "no".\\
\\
Answer:\\
Let's think step by step without any assumptions.\\
\midrule
\textbf{Output:}\\
Grounding check\\
Purpose: Find references for numbers in the reasoning process sequentially\\
Rule: Any ungrounded number makes the check fails\\
\\
"12" is a quantity from the given information that represents the number of doughnuts in a dozen.\\
"10" is a quantity not from the given information, it is ungrounded.\\
"\$2" is a quantity from the given information that represents the price per doughnut.\\
\red{"120" is a quantity not from the given information, it is ungrounded.}\\
Result: Based on the check, there are ungrounded numbers, so the grounding check fails.\\
\\
Reasoning check:\\
Purpose: Check whether the reasoning logic correctly answers the question\\
Rule: Any logic error makes the check fails\\
\\
The reasoning process is correct. We have the number of doughnuts Derrick makes in a day, \\and we have the price per doughnut.\\ Multiplying these two yields the revenue from selling all the doughnuts in a day.\\
Result: Based on the check, the reasoning logic is correct.\\
\\
Calculation check:\\
Purpose: Check whether the calculation process is consistent\\
Rule: Any inconsistent calculation makes the check fails\\
\\
calculation1: \\
equation: 10 * 12, answer: 120\\
10 * 12 = 120 is consistent with the answer, so the calculation is correct.\\
calculation2:\\
equation: \$2 * 120, answer: \$240\\
\$2 * 120 = \$240 is consistent with the answer, so the calculation is correct.\\
Result: Based on the check, the calculation process is consistent, so the calculation check passes.\\
\\
Check results: Ground check fails, Reasoning check passes, Calculation check passes.\\
Rule: Any failed check makes the reasoning incorrect.\\
So the answer is "no".\\
\bottomrule
\end{tabular}
\vspace{1em}
\caption{Failure case: our deductive verification process sometimes treats grounded information as if they were ungrounded. The number 120 is provided in the given information, but the model states that it is ungrounded.}
\label{tab:gsm8k true to false verification}
\end{table*}

\end{document}